%% file: acl_latex.tex
\pdfoutput=1

\documentclass[11pt]{article}
\usepackage{CJKutf8}
\usepackage[final]{acl}
\usepackage{times}
\usepackage{latexsym}
\usepackage{algorithm}
\usepackage{algorithmic}
\usepackage{microtype}
\usepackage{inconsolata}
\usepackage{amsfonts}
\usepackage{amssymb}
\usepackage{amsmath}
\usepackage{graphicx}
\usepackage{bm}
\usepackage{multirow}
\usepackage{multicol}
\usepackage{booktabs}
\usepackage{framed}
\usepackage{enumitem}
\usepackage{xcolor}
\usepackage{color, colortbl}
\usepackage{cleveref}
\usepackage{xspace}
\usepackage{tcolorbox}

\usepackage[T1]{fontenc}

\usepackage[utf8]{inputenc}

\usepackage{microtype}

\usepackage{inconsolata}  

\usepackage{xcolor}
\usepackage{amsthm}
\usepackage{amsmath}
\usepackage[inkscapelatex=false]{svg}
\usepackage{dashrule}
\usepackage{arydshln}
\usepackage{overpic}
\usepackage{marvosym}

\newcommand{\cdashlinelr}[1]{%
  \noalign{\vskip\aboverulesep
           \global\let\@dashdrawstore\adl@draw
           \global\let\adl@draw\adl@drawiv}
  \cdashline{#1}
  \noalign{\global\let\adl@draw\@dashdrawstore
           \vskip\belowrulesep}}

\makeatletter
\def\@fnsymbol#1{}
\makeatother

\newtheorem{insight}{Finding}

\newenvironment{tcolorboxinsight}[1][]{%
    \begin{tcolorbox}[#1]
    \vspace{-0.2cm}
    \begin{insight}
}{%
    \end{insight}
    \vspace{-0.3cm}
    \end{tcolorbox}
}

\title{\textit{ShifCon}: Enhancing Non-Dominant Language Capabilities with a Shift-based Multilingual Contrastive Framework}

\author{
Hengyuan Zhang\textsuperscript{\rm{1 * \dag}}\thanks{*\ This work was done during internship at Microsoft}, 
Chenming Shang\textsuperscript{\rm{1 \dag}}\thanks{\dag\ Equal contribution}, 
Sizhe Wang\textsuperscript{\rm{3}},
\textbf{Dongdong Zhang}\textsuperscript{\rm{2 \Letter}\thanks{\textsuperscript{\Letter}\ Corresponding author}\ }\textbf{,} \\
\textbf{Yiyao Yu}\textsuperscript{\rm{1}}\textbf{,}
\textbf{Feng Yao}\textsuperscript{\rm{4}}\textbf{,}
\textbf{Renliang Sun}\textsuperscript{\rm{5}}\textbf{,}
\textbf{Yujiu Yang}\textsuperscript{\rm{1}\Letter}\textbf{,}
\textbf{Furu Wei}\textsuperscript{\rm{2}} \\
\textsuperscript{1} Tsinghua University \ \
\textsuperscript{2} Microsoft  \ \
\textsuperscript{3} University of Southern California \\
\textsuperscript{4} University of California, San Diego \ \
\textsuperscript{5} University of California, Los Angeles \\
\texttt{\{zhang-hy22,scm22\}@mails.tsinghua.edu.cn} 
}

\begin{document}
\maketitle

\begin{abstract}
\input{abstract/abstract.tex}

\end{abstract}

\section{Introduction}
\label{sec:intro}
\input{intro/intro}

\begin{figure*}[!t]
 \centering
 \centerline{\includegraphics[width=2\columnwidth]{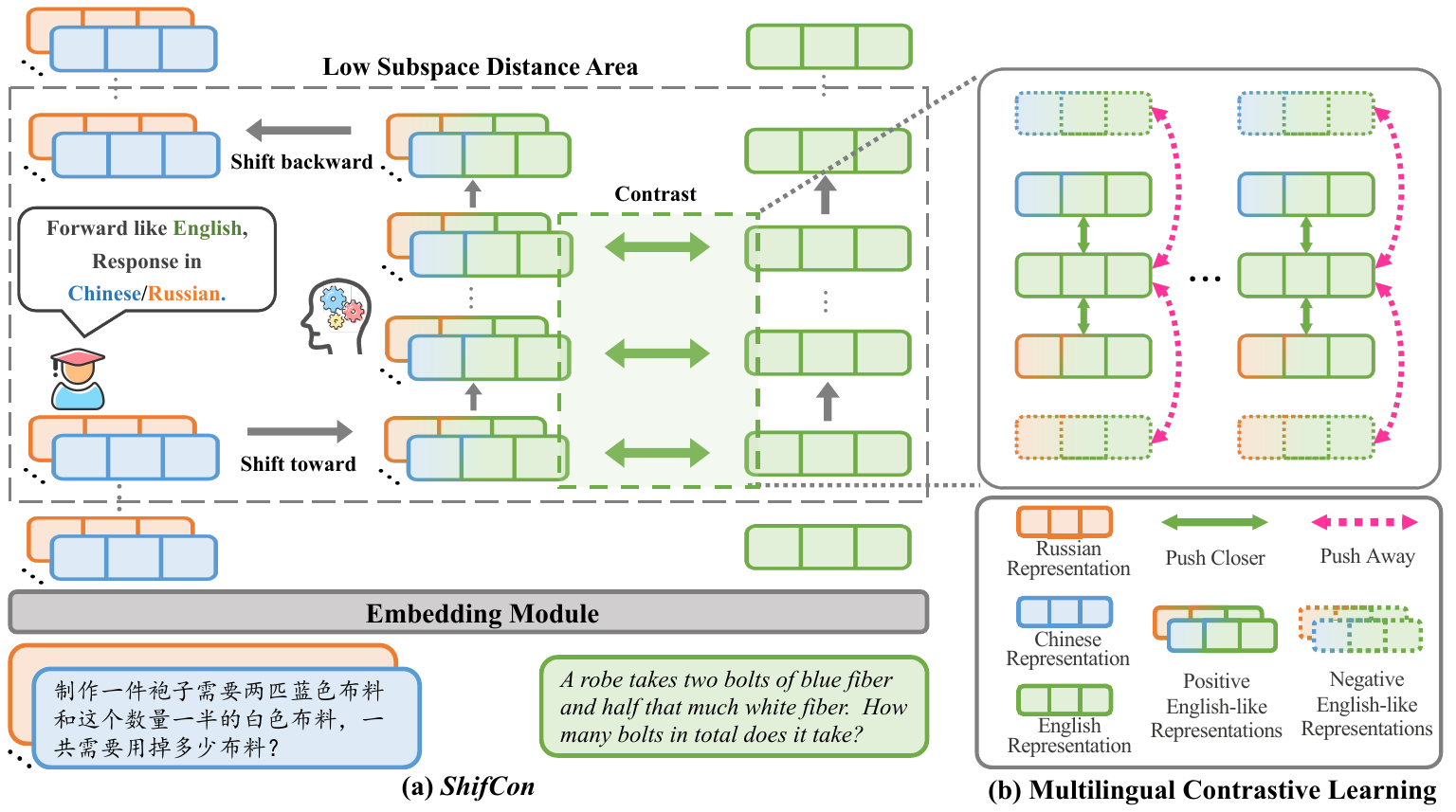}}
    \caption{An illustration of our \textit{ShifCon} framework: (I) We shift non-dominant language representations (e.g., Chinese and Russian) into the dominant language subspace (e.g., English) to obtain their dominant-like representations. (II) Using parallel translation inputs between the non-dominant and dominant languages as positive samples, multilingual contrastive learning pushes non-dominant language's dominant-like representations closer to the dominant language and pushes away them from other representations.}
    \label{fig:overview}
\end{figure*}

\section{The Framework}
Our \textit{ShifCon} (shown in Fig.~\ref{fig:overview}) includes two modules:  
1) Shift Projection (\S~\ref{sec:shift_projection}), which maps the representations of non-dominant language into the dominant language subspace to obtain its dominant-like representations during internal forward process, and then shifts backwards to its native space before generation;
2) Multilingual Contrastive Learning (\S~\ref{sec:multilingual_contrastive}), which further aligns dominant-like representations of non-dominant languages with their dominant language counterparts.

\subsection{Shift Projection}
\label{sec:shift_projection}

\subsubsection{Shift-toward and Shift-backward}
\label{sec:shift_forward_backward}
\input{shift_forward_backward/shift_forward_backward}

\subsubsection{Language Subspace Distance}
\label{sec:language_subspace}
\input{language_subspace/language_subspace}

\input{tabs/main_results}

\subsection{Multilingual Contrastive Learning (MCL)}
\label{sec:multilingual_contrastive}
\input{multilingual_contrastive/multilingual_contrastive}

\section{Experiment}

\subsection{Experiment Settings}
\label{sec:exp_setting}
\input{exp_setting/exp_setting}

\subsection{Performance of \textit{ShifCon}}
\label{sec:results}

\input{results/results.tex}

\subsection{Further Analysis}
\label{sec:further_analysis}
\input{further_analysis/further_analysis.tex}

\section{Related Work}
\label{sec:related_work}
\input{related/related}

\section{Conclusion}
\label{sec:conclusion}
\input{conclusion/conclusion.tex}

\section{Limitations}
\label{sec:limitation}
\input{limitation}

\bibliography{acl_latex}

\newpage
\onecolumn
\appendix
\section{Appendix}
\label{sec:appendix}
\input{appendix.tex}

\end{document}

%% file: abstract/abstract.tex
Although fine-tuning Large Language Models (LLMs) with multilingual data can rapidly enhance the multilingual capabilities of LLMs, they still exhibit a performance gap between the dominant language (e.g., English) and non-dominant ones due to the imbalance of training data across languages.
To further enhance the performance of non-dominant languages, we propose \textit{ShifCon}, a \underline{\textbf{Shif}}t-based multilingual \underline{\textbf{Con}}trastive framework that aligns the internal forward process of other languages toward that of the dominant one. 
Specifically, it shifts the representations of non-dominant languages into the dominant language subspace, allowing them to access relatively rich information encoded in the model parameters.
The enriched representations are then shifted back into their original language subspace before generation.
Moreover, we introduce a subspace distance metric to pinpoint the optimal layer area for shifting representations and employ multilingual contrastive learning to further enhance the alignment of representations within this area.
Experiments demonstrate that our \textit{ShifCon} framework significantly enhances the performance of non-dominant languages, particularly for low-resource ones.
Further analysis offers extra insights to verify the effectiveness of \textit{ShifCon} and propel future research.

%% file: intro/intro.tex
While LLMs have demonstrated strong multilingual capabilities~\citep{lin-etal-2022-shot,achiam2023gpt,anil2023palm2}, a performance gap remains between the dominant language and non-dominant ones, primarily due to the imbalance in training data across languages~\citep{shi2022language, huang2023not,gurgurov2024multilingual}.
A common strategy to mitigate this issue is translating dominant language data into non-dominant languages and applying Multilingual Supervised Fine-Tuning (MSFT) on the resulting multilingual datasets~\citep{chen2023breaking,zhang2023bayling}.

\begin{figure}[!t]
    \centering  \centerline{\includegraphics[width=\columnwidth]{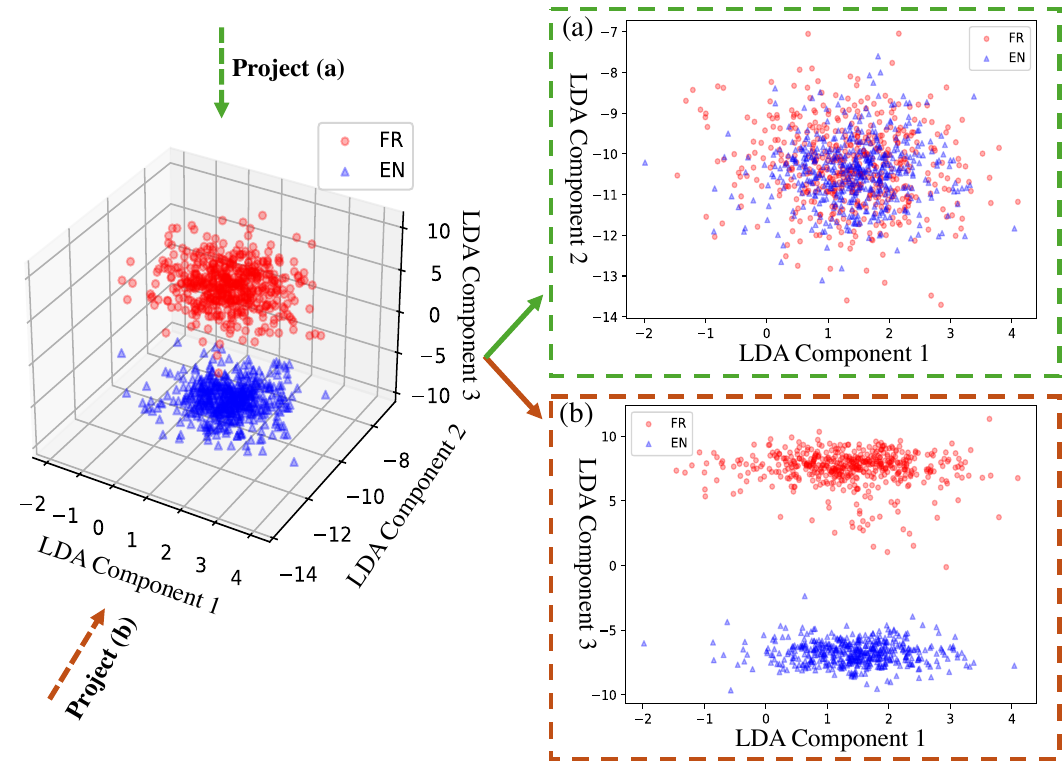}}
    \caption{Two different projections on the sentence representations visualized using LDA. Projection (a) shows the representations are mutually aligned, implying a language-agnostic status, whereas projection (b) illustrates separated representations in distinct spaces, suggesting a language-specific status. The sentence representations are obtained through mean-pooling the hidden states from the 15th layer of $\text{Llama-2}_\text{7B}$.}
    \label{fig:intro}
\end{figure}
While MSFT provides initial capabilities for non-dominant languages, two key challenges limit further progress: 
1) annotating high-quality data for non-dominant languages is expensive, even for the dominant language that serves as the source for translation~\citep{kholodna2024llms};
2) translation errors often lead to error propagation in subsequent procedures~\citep{agrawal-etal-2024-translation}, thus requiring extensive verification to ensure data quality.
As a result, high-quality data for non-dominant languages is limited in scale, which restricts the effectiveness of MSFT. This raises an important question: \textit{Can we improve the performance of non-dominant languages with limited MSFT data?}

Considering this external limitation, previous work has delved into exploring internal representation alignment to improve performance~\citep{yoon-etal-2024-langbridge,li-etal-2024-improving-context}.
A growing consensus indicates that it is the language-agnostic representations, which are exhibited in the middle layer of the model, facilitating this enhancement~\citep{kojima-etal-2024-multilingual,tang-etal-2024-language}.
Beyond those efforts, we consider that the representations, even in the middle layer, still retain language-specific information.
Specifically, by visualizing sentence representations of translation pairs using linear discriminant analysis (LDA) in Fig.~\ref{fig:intro}, we observe representations under projection (a) in the middle layer are mapped closely together (e.g., the 15th layer of $\text{Llama-2}_\text{7B}$ out of 32 layers), suggesting a language-agnostic status, consistent with findings in prior research. 
However, in projection (b), we find that different languages occupy distinct subspaces across layers, indicating that language-specific information is consistently encoded within the representations (See Appendix~\ref{sec:langs_subspace_across_layers} for complete results across all languages, layers, and models). 
This information enables the model to differentiate between languages.
Moreover, we consider the superior performance of dominant languages is due to their representations being able to access more information during the internal forward process. 
This is because dominant language data predominates during pre-training, so much of the model's knowledge is encoded in the dominant language format, which is more easily accessible through its representations~\citep{kassner-etal-2021-multilingual,yin-etal-2022-geomlama,zhao-etal-2024-tracing}.

Based on these findings, we propose a \underline{\textbf{Shif}}t-based multilingual \underline{\textbf{Con}}trastive framework (\textit{ShifCon}) to boost the performance of non-dominant language.
It includes shift-toward and shift-backward projections, as well as multilingual contrastive learning (MCL). 
The shift-toward process maps non-dominant language representations into the dominant language subspace to obtain their dominant-like representations, allowing them to access more information encoded in the model, similar to how the dominant language operates.
As language-specific information is crucial for generating outputs in the target language~\citep{li-murray-2023-zero,xu-etal-2023-language-representation,tang-etal-2024-language}, a shift-backward process is needed to project the enriched dominant-like representations back into the original non-dominant language subspace before generation.
During this process, a  subspace distance metric is introduced to pinpoint the optimal layer area for shifting representations.
Moreover, our analysis reveals that even after shifting,  the alignment between non-dominant language's dominant-like representations and their dominant language counterparts remains insufficient.
Therefore, we further apply multilingual contrastive learning to enhance their alignment.

To summarize, our contributions are as follows:

\vspace{0.5em}
\quad \textbf{1)} We present \textit{ShifCon} framework, designed to boost the performance of non-dominant languages by aligning their internal forward process with that of the dominant language.
We also define a subspace distance metric to pinpoint the optimal layer area for implementing shift projection.

\vspace{0.5em}
\quad \textbf{2)} Extensive experiments validate the efficacy of \textit{ShifCon} across diverse tasks and model scales, e.g., a 18.9\% improvement on MGSM for low-resource languages in $\text{Llama-2}_\text{7B}$.
Further analysis confirms the effectiveness of the identified layer area for shift projection using subspace distance metric. 
The improved alignment between dominant-like representations and their dominant counterparts enhances overall performance.

\vspace{0.5em}
\quad \textbf{3)} Moreover, we give the speculation that 30\% of model layers with the lowest distance are likely focused on information aggregation and show that directly applying MCL to original representations may compromise the language-specific information within representations, which impedes the model’s ability to generate in that language.

%% file: shift_forward_backward/shift_forward_backward.tex
To obtain the dominant-like representations for non-dominant languages, thereby enabling them to access more information encoded in the model parameters during the internal forward process, our shift-toward module maps non-dominant language representations into dominant language subspace.

Specifically, given an input query in a non-dominant language $l$, the shift-toward process can be formulated as follows:
\begin{equation}
\label{eq:shifted_representation}
    \bm{\tilde{h}}^{L_\text{to}}_l = \bm{h}^{L_\text{to}}_l -  \bm{v}^{L_\text{to}}_l + \bm{v}^{L_\text{to}}_{d}\ (1\leq L_{\text{to}}  \textless {L})
\end{equation}
\noindent where $L_\text{to}$ is the layer we shift the representation toward, $\bm{h}^{L_\text{to}}_l \in \mathbb{R}^{n \times d}$ denotes $L_\text{to}$-th layer hidden states of the input query in language $l$, where $n$ is the number of tokens in the input query, $d$ is the hidden dimension of the LLM. 
$\bm{v}^{L_{\text{to}}}_l\in\mathbb{R}^d$ and $\bm{v}^{L_{\text{to}}}_{d}\in\mathbb{R}^d$ are the $L_{\text{to}}$-th layer language vectors for the non-dominant language $l$ and the dominant language, respectively.\footnote{We utilize language vectors in the shift projection process, as it has been demonstrated to be an effective approach for language space mapping~\citep{libovicky-etal-2020-language,xu-etal-2023-language-representation,tang-etal-2024-language}.}
To compute the language vectors across all layers for each language $l$, a set of sentences in that language is fed into the LLM. 
From the $i$-th layer of the LLM, sentence vectors are obtained by pooling the token representations\footnote{We explore different pooling methods in Appendix~\ref{sec:appendix_pooling_methods}.} within the sentence. These sentence vectors are then averaged to produce $\bm{v}_l^{i}\in\mathbb{R}^d$. 
In this way, we gather a set of vectors $\mathcal{V}_l=[\bm{v}_l^{1},\bm{v}_l^{2},...,\bm{v}_l^{L}]$, where ${L}$ denotes the number of layers in the LLM. 
The obtained dominant-like representations of non-dominant language are then fed to the succeeding layers to access relatively rich information encoded in the model parameters.

Since language-specific information is crucial for models to generate answers in that language, we shift dominant-like representations of the non-dominant language back to its native subspace at the $L_{bk}$-th layer before generation:
\begin{equation}
    \bm{h'}^{L_{\text{bk}}}_l = \bm{\tilde{h}}^{L_{\text{bk}}}_l - \bm{v}^{L_{\text{bk}}}_{\text{d}} + \bm{v}^{L_{\text{bk}}}_{l}\ ({L_{\text{to}}}\textless {L_{\text{bk}}}\leq {L})
\end{equation}
\noindent where ${L_{\text{bk}}}$ is the layer we shift the representation backward, $\bm{\tilde{h}}^{L_{\text{bk}}}_l$ represent the ${L_{\text{bk}}}$-th layer hidden states of non-diminant language $l$. 
$\bm{\tilde{h}}^{L_{\text{bk}}}_l$ are dominant-like representations because of the shift-toward projection. 
They are shifted back into their original subspace, resulting in $\bm{h'}^{L_{\text{bk}}}_l$. The representations, now containing language-specific information of $l$, are then fed into the subsequent layers to produce responses in language $l$.

%% file: language_subspace/language_subspace.tex
It is crucial to establish an effective criterion for determining the 
optimal layer area for conducting shift projection procedure.
A practical solution is to select layers where the subspace of non-dominant language's dominant-like representations\footnote{We term the subspace of non-dominant language's dominant-like representations as ``dominant-like subspace''.} aligns well with the subspace of the dominant language counterparts, as greater alignment indicates they can be more similar in the internal forward process.

Therefore, we introduce a subspace distance metric to measure the alignment between their subspaces, where smaller distances indicating stronger alignment.
Specifically, for the language $A$, we define an affine subspace $\mathcal{S^A}$ using the language's mean representation $\bm{\mu}_A \in \mathbb{R}^d$ along with $k_{A}$ principal directions of maximal variance in the language, defined by an orthonormal basis $\bm{V}_A \in \mathbb{R}^{d \times k_{A}}$.
We consider this basis with $k_{A}$ directions can best describe the language-specific information of language $A$.
To identify this subspace, we use $\bm{X}_A \in \mathbb{R}^{n \times d}$ to obtain $\bm{\mu}_A$ and employ singular value decomposition (SVD) on the $\bm{X}_A$ to obtain $\bm{V}_A$, which is selected from the top-$k_{A}$ singular value by $\bm{\Sigma}_A \in \mathbb{R}^{k_{A} \times k_{A}}$.
Here, $\bm{X}_A$ donates $n$ contextualized token representations with $d$ dimensionality in language $A$ from the desired layer.
We select the subspace dimensionality $k$ such that the subspace accounted for 90\% of the total variance in the language.\footnote{See more details of computing process in Appendix~\ref{sec:detail_language_subspace_distance}.}

Due to the varying dimensionality $k$ of $\bm{V}$ across different languages, we adopt a Riemannian distance metric that measures distances between positive definite matrices~\citep{bonnabel-sepulchre-2009-riemannian,chang-etal-2022-geometry} to quantify the distance between dominant-like subspace $\mathcal{S^{D'}}$ and corresponding dominant language subspace $\mathcal{S^D}$:\footnote{After applying shift projection, the centroids of two subspaces will coincide, causing $||\bm{\mu}_{D'}-\bm{\mu}_{D}||_2 = 0$.}
\begin{equation}
\small
    \textrm{Dist}(\mathcal{S^{D'}}, \mathcal{S^D}) = \sqrt{\sum_{i=1}^d \log^2(\lambda_i)} + ||\bm{\mu}_{D'}-\bm{\mu}_{D}||_2 
\end{equation}

\noindent where $\lambda_i$ is the $i$-th positive real eigenvalue of $\bm{K}_{{D'}}^{-1} \bm{K}_{D}$. Here $\bm{K}_{{D}} \in \mathbb{R}^{d \times d}$ can be calculated from the SVD of the right singular matrices $\bm{V}_{{D}}$:
\begin{equation}
    \bm{K}_{{D}} = \frac{1}{n-1} \bm{V}_{{D}} \bm{\Sigma}_{{D}}^2 \bm{V}_{{D}}^T
\end{equation}

\begin{figure}[!t]
    \centering
    \begin{overpic}[width=\columnwidth]{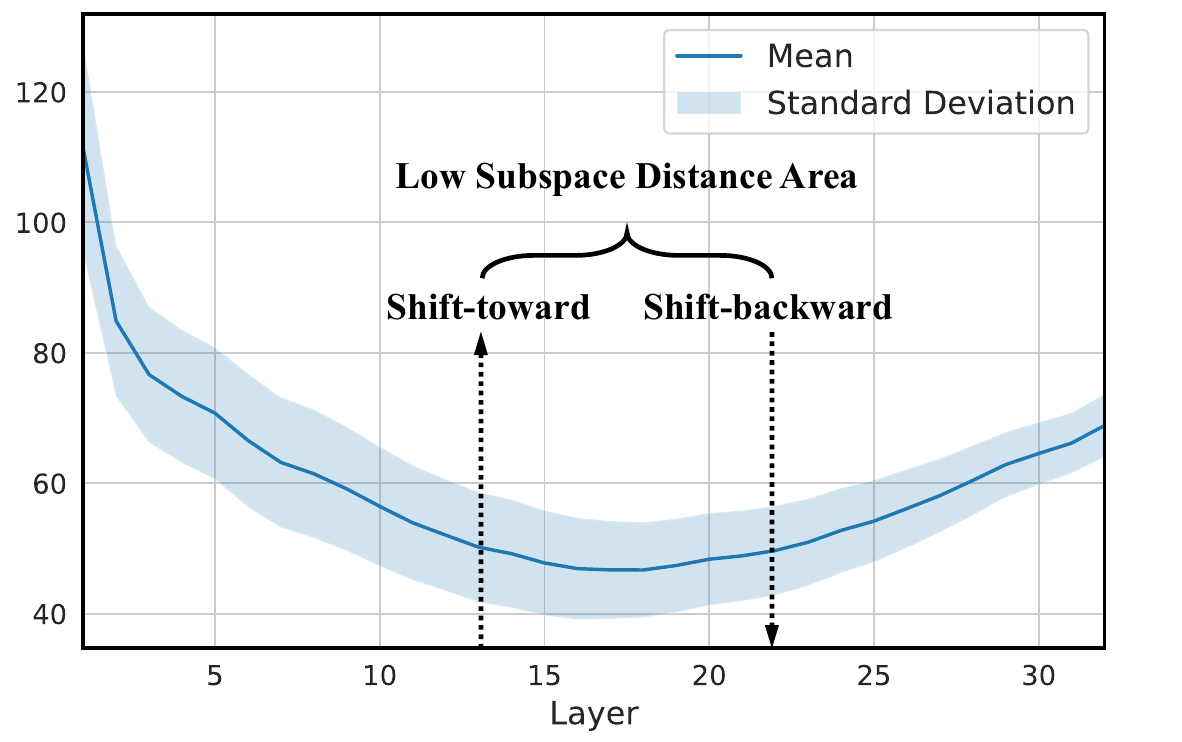}\end{overpic}
    \caption{The distance of dominant-like subspace $\mathcal{S^{D'}}$ and corresponding dominant language subspace $\mathcal{S^D}$ in the $\text{XGLM}_\text{7.5B}$ using 1k FLORES samples per language. Its low subspace distance area, [13, 22], identified by $\beta$=30\% (Finding~\ref{insight:2}), indicating shifting towards in the 13th layer and backward in the 22nd layer.}
    \label{fig:method-low-sub-dist}
\end{figure}

We present the distance results of the $\text{XGLM}_\text{7.5B}$ in Fig.~\ref{fig:method-low-sub-dist}. 
We observe that the subspace distances in the middle layers are minimal, while the distances on the sides are larger with steep slopes. 
This observation suggests that the middle layers in the model achieves superior alignment between dominant-like representations and their dominant language counterparts, enabling them access richer information analogous to dominant language representations, rendering it suitable for shift projection.

To precisely identify these layers, we propose a simple method of sorting the distances in ascending order and selecting the top-$\beta$\footnote{We test $\beta$ from 0\% to 100\%, choosing $\lceil N\times\beta \rceil$ layers to define the low subspace distance area. \( \lceil \cdot \rceil \) is ceiling function.} layers with the smallest distances to establish the \textit{low subspace distance area}.
We find that the layers within the low subspace distance area are contiguous across models of different families and scales, making them ideally suited for shift projection.

%% file: tabs/main_results.tex
\begin{table*}[!th]
    \renewcommand\arraystretch{1.1}
    
    \centering
    
    \setlength\tabcolsep{4pt}
    \fontsize{10}{10}\selectfont 
    
    \begin{tabular}{lccccccccccccccccc}
        \toprule[1.2pt]
        & \multicolumn{8}{c}{\textbf{Generation}} & \textbf{}  & \multicolumn{8}{c}{\textbf{Classification}}   \\ \addlinespace[2pt]
        \cline{2-9} \cline{11-18} \addlinespace[2pt]
        
                                            & \multicolumn{2}{c}{\textbf{MGSM}} & \textbf{} & \multicolumn{2}{c}{\textbf{\begin{tabular}[c]{@{}c@{}}FLORES\\  (en-xx)\end{tabular}}} & \textbf{} & \multicolumn{2}{c}{\textbf{\begin{tabular}[c]{@{}c@{}}FLORES\\  (xx-en)\end{tabular}}} & \textbf{} & \multicolumn{2}{c}{\textbf{XCOPA}} & \textbf{} & \multicolumn{2}{c}{\textbf{XNLI}} & \textbf{} & \multicolumn{2}{c}{\textbf{XStoryCloze}}   \\ \addlinespace[2pt]
                                            \cline{2-3} \cline{5-6} \cline{8-9} \cline{11-12} \cline{14-15} \cline{17-18} \addlinespace[2pt]
                          & \textbf{High}   & \textbf{Low}    & \textbf{} & \textbf{High}        & \textbf{Low}         & \textbf{} & \textbf{High}        & \textbf{Low}         & \textbf{} & \textbf{High}    & \textbf{Low}    & \textbf{} & \textbf{High}   & \textbf{Low}    & \textbf{} & \textbf{High}       & \textbf{Low}       \\ \addlinespace[2pt] \hline \addlinespace[2pt] 
        $\text{Llama-2}_\text{7B}$       & 35.2            & 5.1             &           & 33.5                 & 15.9                 &           & 39.8                 & 21.4                 &           & 63.2             & 49.7            &           & 45.2            & 35.2            &           & 74.7                & 56.6               \\
         \ \ \ \ \ \ \ \ \ +MSFT              & 44.9            & 29.5            &           & 34.7                 & 18.4                 &           & 40.4                 & 24.7                 &           & 64.2             & 52.0            &           & 46.4            & 37.6            &           & 75.3                & 58.7               \\
          \ \ \ \ \ \ \ \ \ +AFP              & 46.3            & 31.7            &           & 35.2                 & 19.1                 &           & 41.0                 & 25.3                 &           & 65.0             & 52.8            &           & 46.8            & 38.7            &           & 76.0                & 59.8               \\
        \ \ \ \ \ \ \ \ \ +\textit{ShifCon} & \textbf{48.2}   & \textbf{35.1}   & \textbf{} & \textbf{35.6}        & \textbf{19.7}        & \textbf{} & \textbf{41.8}        & \textbf{26.4}        & \textbf{} & \textbf{65.5}    & \textbf{53.5}   & \textbf{} & \textbf{47.2}   & \textbf{40.1}   & \textbf{} & \textbf{76.6}       & \textbf{60.8}       \\ \addlinespace[2pt] \hdashline[1pt/1pt] \addlinespace[2pt]
        $\text{XGLM}_\text{7.5B}$    & 4.0             & 1.9             &           & 32.2                 & 31.5                 &           & 41.2                 & 35.8                 &           & 63.8             & 57.3            &           & 44.5            & 41.4            &           & 65.2                & 58.4               \\
         \ \ \ \ \ \ \ \ \ +MSFT              & 10.6            & 7.0             &           & 33.5                 & 32.8                 &           & 42.3                 & 37.3                 &           & 64.9             & 58.3            &           & 45.9            & 42.3            &           & 66.7                & 60.1               \\
         \ \ \ \ \ \ \ \ \ +AFP              & 12.1            & 9.6             &           & 34.0                 & 33.3                 &           & 43.2                 & 37.7                 &           & 65.7             & 58.9            &           & 47.0           & 43.3            &           & 67.4                & 60.9               \\
         \ \ \ \ \ \ \ \ \ +\textit{ShifCon}  & \textbf{13.7}   & \textbf{11.7}   & \textbf{} & \textbf{34.5}        & \textbf{34.1}        & \textbf{} & \textbf{43.7}        & \textbf{38.5}        & \textbf{} & \textbf{66.8}    & \textbf{60.1}   & \textbf{} & \textbf{48.6}   & \textbf{44.3}   & \textbf{} & \textbf{68.1}       & \textbf{62.2}      \\ \addlinespace[2pt] \hdashline[1pt/1pt] \addlinespace[2pt]
        $\text{BLOOM}_\text{7.1B}$   & 13.2            & 3.7             &           & 41.4                 & 24.3                 &           & 45.7                 & 30.7                 &           & 57.7             & 52.1            &           & 42.4            & 36.6            &           & 67.3                & 58.1               \\
         \ \ \ \ \ \ \ \ \ +MSFT              & 21.9            & 12.5            &           & 42.3                 & 25.9                 &           & 46.5                 & 33.1                 &           & 59.2             & 53.9            &           & 44.0            & 38.9            &           & 68.6                & 59.8               \\
         \ \ \ \ \ \ \ \ \ +AFP              & 22.9            & 15.7            &           & 43.0                 & 26.6                 &           & 47.0                 & 33.6                 &           & 59.9             & 54.8            &           & 44.9            & 39.9            &           & 68.9                & 60.2               \\
         \ \ \ \ \ \ \ \ \ +\textit{ShifCon}  & \textbf{24.5}   & \textbf{18.8}   & \textbf{} & \textbf{43.4}        & \textbf{27.2}        & \textbf{} & \textbf{47.2}        & \textbf{34.5}        & \textbf{} & \textbf{60.3}    & \textbf{56.3}   & \textbf{} & \textbf{45.5}   & \textbf{40.8}   & \textbf{} & \textbf{69.5}       & \textbf{60.9}      \\ \specialrule{0em}{0pt}{0pt}   \bottomrule[1.2pt]  
        \end{tabular}
        \caption{The average results of high- and low-resource languages across five tasks within three distinct model families. Detailed results for each language can be found in Appendix~\ref{sec:detailed_results}. ``en-xx'' denotes translation from English to another language, while ``xx-en'' indicates translation from another language to English. Base model, e.g., $\text{Llama-2}_\text{7B}$, indicates fine-tuning solely with English data.}
        \label{tab:main_results}
        \end{table*}

%% file: multilingual_contrastive/multilingual_contrastive.tex
However, as shown in Fig.~\ref{fig:method-low-sub-dist}, some subspace distance still remains, even in the low subspace distance area (e.g., $\text{XGLM}_\text{7.5B}$'s 16th layer still exhibits a subspace distance of about 47), which requires further alignment to reduce.
To address this, we employ multilingual contrastive learning to achieve a more refined alignment. 
We use translation pairs from dominant and non-dominant languages as positive pairs, pulling the dominant-like representations of non-dominant language closer to their dominant language counterparts. 
While the dominant-like representations of other sentences in the same batch serve as negative samples.

Formally, given a mini-batch of translation pairs from non-dominant and dominant languages $\{(s_{l}^{i}, s_{d}^{i})\}_{i=1}^{N}$, the Multilingual Contrastive Learning (MCL) loss at the $t$-th layer is:
\begin{equation}
\small
\begin{split}
\bm{\tilde{e}}_{l}^{i} &= g(\left[\bm{\tilde{h}}_{l}^{t}\right]^{i}); \quad \quad \bm{e}_{d}^{i} = g(\left[\bm{h}_d^t\right]^{i}) \\
\mathcal{L}^{t}_{\textit{MCL}}(\theta) &= \sum^{N}_{i=1} -{\text{log} \frac{\text{exp}(\text{sim}(\bm{\tilde{e}}^{i}_{l},\bm{e}^{i}_{d})/\tau)}{\sum_j{\text{exp}(\text{sim}(\bm{\tilde{e}}^{i}_{l},\bm{e}^{j}_{d})/\tau)}}}
\end{split}
\end{equation}

\noindent where $g(\cdot)$ is the pooling method used to obtain sentence representations, $\left[\bm{\tilde{h}}_{l}^{t}\right]^{i}$ denotes the $t$-th layer dominant-like representations of $s_{l}^{i}$, $\left[\bm{h}_d^t\right]^{i}$ is the $t$-th layer representations of $s_{d}^{i}$, and $\text{sim}(,)$ is cosine similarity function.
$\tau$ is a temperature hyperparameter. 
MCL is performed on the layers between $[L_{\text{to}}, L_{\text{bk}})$ to achieve better alignment, resulting in the total MCL loss: $\mathcal{L}_{\textit{MCL}} = \sum^{L_{\text{bk}}-1}_{t=L_{\text{to}}}\mathcal{L}^{t}_{\textit{MCL}}$.

We illustrate the process of MCL in Fig.~\ref{fig:overview} (b) and train our \textit{ShifCon} using the following loss:
\begin{equation}
\mathcal{L}_{\textit{ShifCon}}(\theta) = \mathcal{L}_{\textit{MSFT}}(\theta) + \alpha \mathcal{L}_{\textit{MCL}}(\theta)
\label{eq:total_loss}
\end{equation}

\noindent where $\mathcal{L}_{\textit{MSFT}}$ denotes the loss of MSFT, computed through autoregressive language modeling on the multilingual dataset, and $\alpha \in \mathbb{R}_{+}$ is a hyper-parameter to balance these two losses. 
It is important to note that when computing $\mathcal{L}_{\textit{MSFT}}$ for non-dominant language samples, their dominant-like representations are used during the internal forward process instead of their original ones.\footnote{In this work, we introduce a new strategy to obtain better language vectors for shift projection in the training phase. The details are illustrated in Appendix~\ref{sec:appendix_new_strategy}.}

%% file: exp_setting/exp_setting.tex
\paragraph{Evaluation Tasks} 
We conduct evaluations on a variety of multilingual benchmarks, covering both generation and classification tasks.
\textbf{1)} For generation tasks, we consider FLORES~\citep{nllb2022}, a benchmark for machine translation, and MGSM~\citep{shilanguage}, a multilingual math reasoning task. 
\textbf{2)} For classification tasks, we utilize XNLI~\citep{conneau-etal-2018-xnli}, XCOPA~\citep{ponti-etal-2020-xcopa}, and XStoryCloze~\citep{lin-etal-2022-shot}, which are widely used generic reasoning datasets.

For the evaluation of MGSM, we utilize MGSM8KInstruct~\citep{chen2023breaking} as the training set, which translates the GSM8K into nine non-English languages. 
For the evaluation of the other tasks, we follow~\citet{li-etal-2024-improving-context} and utilize Bactrian-X~\citep{li2023bactrian}, which has been translated into 52 languages from Alpaca~\citep{alpaca} and Dolly~\citep{DatabricksBlog2023dolly}, as the training set. 
See Appendix~\ref{sec:datasets_and_evaluation} for more details about the datasets we used in the experiment.

\paragraph{Metrics}
For MGSM, we implement a rule-based extraction strategy~\citep{chen2023breaking} to derive accuracy results in a zero-shot manner.
We utilize the evaluation framework introduced by~\citet{zhang-etal-2024-impact} for assessing the other benchmarks in a 4-shot manner. 
Specifically, we assess the performance on the FLORES dataset using ChrF++~\citep{popovic-2017-chrf} score, while the performance on the other datasets is evaluated based on rank classification accuracy.\footnote{The scoring function averages per-token logarithmic probabilities, excluding shared prefixes. The candidate with the highest score is chosen as the prediction.}

\paragraph{Training Setup}
We incorporate LLMs from different families, such as Llama~\citep{touvron2023llama2openfoundation}, BLOOM~\citep{scao2022bloom}, and XGLM~\citep{lin-etal-2022-shot}, in our experiments. 
We utilize \textbf{\emph{English}} as the dominant language in these three model families, as its data predominates in their corresponding pre-training corpus.
The models trained using \textbf{MSFT} and the state-of-the-art alignment framework \textbf{AFP}~\citep{li-etal-2024-improving-context}, serve as the baseline for comparison.
Since both MGSM8KInstruct and Bactrian-X are constructed through translation, we directly extract the instruction content from their respective datasets to acquire the translation pairs for MCL.
The details of model information and training settings can be found in Appendix~\ref{sec:implementation_details}.

%% file: results/results.tex
\input{main_results/main_results}

%% file: main_results/main_results.tex
\begin{figure}[!t]
    \centering  \centerline{\includegraphics[width=\columnwidth]{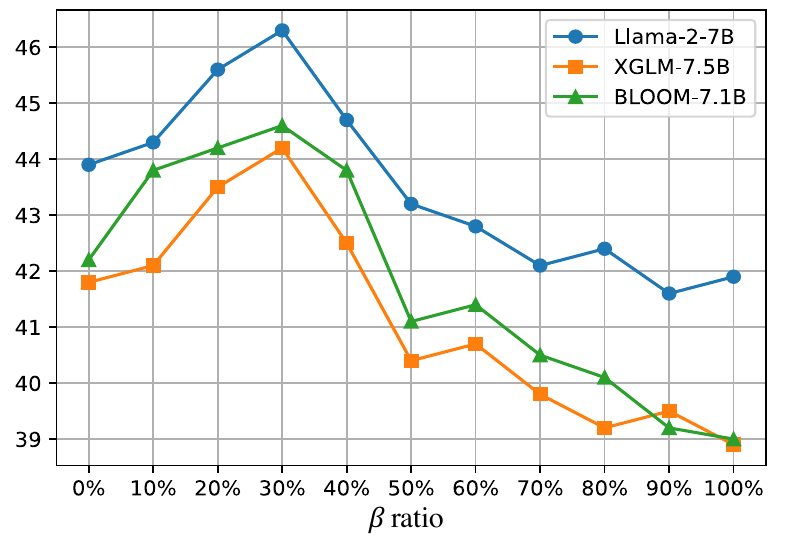}}
    \caption{The average results of all benchmarks across different $\beta$ ratios in three distinct family models.}
    \label{fig:layer_ratio}
\end{figure}

We categorize the experimental languages into high- and low-resource languages based on their data ratios in the LLM pre-training corpus, and report their average results across different tasks in Table~\ref{tab:main_results}.
As shown in Table~\ref{tab:main_results}, despite the initial capabilities provided by MSFT for non-dominant languages, our \textit{ShifCon} consistently further boosts their performance.
Specifically, for $\text{XGLM}_\text{7.5B}$, our \textit{ShifCon} improves performance by 2.1\% for the high-resource languages on XCOPA and a more substantial improvement of 3.5\% for the low-resource languages. 
Moreover, we observe that the enhancement of multilingual understanding also facilitates generation. 
For example, \textit{ShifCon} exhibits an improvement of 7.3\% on high-resource languages on MGSM and a more significant improvement of 18.9\% on low-resource languages.
Based on these observations, we conclude that: \textit{ShifCon} \emph{improves the performance of non-dominant languages, especially for low-resource languages.}

%% file: further_analysis/further_analysis.tex
\paragraph{Suitable $\beta$ for Shift Projection} 
\label{sec:layer_ratio_for_shift}
\input{layer_ratio_for_shift/layer_ratio_for_shift}

\paragraph{Performance of \textit{ShifCon} across Different Scales}
\label{sec:different_sizes_families}
\input{different_sizes_families/different_sizes_families}

\paragraph{Impact of Shift Projection and MCL} 
\label{sec:ablation_study}
\input{ablation_study/ablation_study}

\paragraph{Low Subspace Distance Area}
\label{sec:low_subspace area}
\input{low_subspace_area/low_subspace_area}

\paragraph{Effectiveness of Subspace Distance Area and Metric}
\label{sec:subspace_distance_effectiveness}
\input{lsd_effectiveness/lsd_effectiveness.tex}

%% file: layer_ratio_for_shift/layer_ratio_for_shift.tex
\input{tabs/different_scales}

\begin{figure*}[!t]
    \centering  \centerline{\includegraphics[width=2\columnwidth]{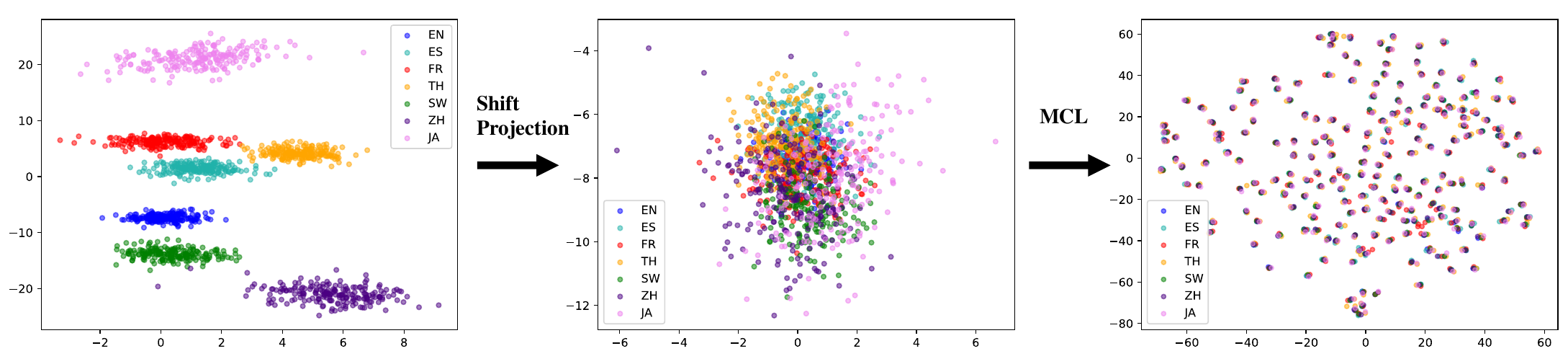}}
    \vspace{-0.2cm}
    \caption{Pooled sentence representations obtained with 300 FLORES samples per language from 15th layer of $\text{Llama-2}_\text{7B}$ after utilizing shift projection and MCL modules. Visualization is based on LDA components 1 and 3.}
    \label{fig:distribution_shift}
\end{figure*}

We conduct extra experiments to determine the number of layers for non-dominant languages to perform in their dominant-like representation during the internal forward process.
In Fig.~\ref{fig:layer_ratio}, the average performance of all benchmarks across three model families is shown for various selection ratios $\beta$ (as defined in \S~\ref{sec:shift_projection}), ranging from 0\% to 100\%. 
The results indicate a trend of initially increasing, peaking at a value of 30\%, and subsequently declining. 
Similar trends can be observed in three models of different families.
Therefore, we set $\beta$ to 30\% by default to obtain the \emph{low subspace distance area} in our \textit{ShifCon} framework and give the following speculation:
\begin{tcolorboxinsight}
    $N \times 30\%$ of layers with lowest subspace distance are likely focused on information aggregation, making them suitable for non-dominant languages to forward in dominant-like representations.
    \label{insight:2}
\end{tcolorboxinsight}

\noindent Where $N$ denotes the number of layers in the model, and this speculation also aligns with the findings observed by~\citet{zhang-etal-2024-balancing}.

%% file: tabs/different_scales.tex
\begin{table}[!t]
\renewcommand\arraystretch{1.1}

\centering

\setlength\tabcolsep{2.5pt}
\fontsize{9}{10}\selectfont 

\begin{tabular}{lcccccccc}
\toprule[1.2pt]
                  & \multicolumn{2}{c}{\textbf{XCOPA}} & \textbf{} & \multicolumn{2}{c}{\textbf{XNLI}} & \textbf{} & \multicolumn{2}{c}{\textbf{XStoryCloze}} \\  \cline{2-3} \cline{5-6} \cline{8-9} \addlinespace[2pt]
                  & \textbf{High}    & \textbf{Low}    & \textbf{} & \textbf{High}   & \textbf{Low}    & \textbf{} & \textbf{High}       & \textbf{Low}       \\ \addlinespace[2pt] \hline \addlinespace[2pt] 
$\text{XGLM}_\text{564M}$         & 54.3             & 51.1            &           & 37.6            & 35.2            &           & 56.1                & 53.0               \\
 \ \ \ \ \ \ \ \ \ +MSFT              & 56.5             & 52.7            &           & 40.4            & 37.8            &           & 57.5                & 55.4               \\
 \ \ \ \ \ \ \ \ \ +AFP               & 57.3             & 53.9            &           & 41.5            & 39.1            &           & 58.0                & 56.6               \\
 \ \ \ \ \ \ \ \ \ +\textit{ShifCon} & \textbf{58.4}    & \textbf{55.8}   & \textbf{} & \textbf{42.6}   & \textbf{40.5}   & \textbf{} & \textbf{59.8}       & \textbf{58.1}      \\ \addlinespace[2pt] \hdashline[1pt/1pt] \addlinespace[2pt]
$\text{XGLM}_\text{2.9B}$    & 61.5             & 54.9            &           & 41.8            & 37.6            &           & 61.7                & 54.9               \\
 \ \ \ \ \ \ \ \ \ +MSFT              & 63.4             & 57.2            &           & 44.6            & 40.5            &           & 64.1                & 57.6               \\
 \ \ \ \ \ \ \ \ \ +AFP               & 64.0             & 58.4            &           & 45.2            & 41.4            &           & 65.3                & 58.8               \\
 \ \ \ \ \ \ \ \ \ +\textit{ShifCon}  & \textbf{65.5}    & \textbf{59.8}   & \textbf{} & \textbf{46.8}   & \textbf{43.3}   & \textbf{} & \textbf{66.5}       & \textbf{60.4}      \\ \addlinespace[2pt] \hdashline[1pt/1pt] \addlinespace[2pt]
$\text{BLOOM}_\text{560M}$   & 53.8             & 51.2            &           & 39.8            & 34.2            &           & 60.3                & 54.2               \\
 \ \ \ \ \ \ \ \ \ +MSFT              & 55.1             & 52.3            &           & 41.7            & 35.4            &           & 62.2                & 54.1               \\
 \ \ \ \ \ \ \ \ \ +AFP               & 55.8             & 53.2            &           & 42.6            & 36.6            &           & 62.8                & 55.3               \\
 \ \ \ \ \ \ \ \ \ +\textit{ShifCon}  & \textbf{56.7}    & \textbf{54.8}   & \textbf{} & \textbf{43.5}   & \textbf{38.2}   & \textbf{} & \textbf{63.6}       & \textbf{56.8}      \\ \addlinespace[2pt] \hdashline[1pt/1pt] \addlinespace[2pt]
$\text{BLOOM}_\text{1.7B}$       & 55.4             & 51.7            &           & 41.5            & 35.3            &           & 62.4                & 54.8               \\
 \ \ \ \ \ \ \ \ \ +MSFT              & 56.9             & 53.4            &           & 43.2            & 36.3            &           & 64.6                & 56.3               \\
 \ \ \ \ \ \ \ \ \ +AFP               & 57.8             & 54.5            &           & 44.0            & 37.3            &           & 65.2                & 57.6               \\
 \ \ \ \ \ \ \ \ \ +\textit{ShifCon}  & \textbf{58.7}    & \textbf{55.8}   & \textbf{} & \textbf{44.8}   & \textbf{38.9}   & \textbf{} & \textbf{66.8}       & \textbf{59.2}      \\  \addlinespace[2pt] \hdashline[1pt/1pt] \addlinespace[2pt]
 $\text{Llama-3}_\text{8B}$         & 68.6             & 54.3            &           & 50.6            & 41.5            &           & 78.5                & 63.9               \\
                  \ \ \ \ \ \ \ \ \ +MSFT              & 69.0             & 55.1            &           & 51.1            & 42.4            &           & 78.8                & 64.7               \\
                  \ \ \ \ \ \ \ \ \ +AFP               & 69.3             & 56.0            &           & 51.3            & 43.1            &           & 79.1                & 65.6               \\
                  \ \ \ \ \ \ \ \ \ +\textit{ShifCon} & \textbf{69.7}    & \textbf{56.9}   & \textbf{} & \textbf{51.6}   & \textbf{44.2}   & \textbf{} & \textbf{79.5}       & \textbf{66.4}      \\
 \bottomrule[1.2pt] 
\end{tabular}
\caption{The average performance of high- and low-resource languages across three classification tasks under model of different scales and families. Base model indicates fine-tuning solely with English data.}
\label{tab:different_scales}
\vspace{-0.2cm}
\end{table}

%% file: different_sizes_families/different_sizes_families.tex
\input{tabs/ablation_study}

Having verified the effectiveness of our \textit{ShifCon} across different model families, we further assess its generalization on different model scales across three classification datasets.
In the BLOOM family models, experiments are conducted at scales of 560M and 1.7B. For the XGLM family models, we utilize 564M and 2.9B scales, and for the Llama family model, we employ the $\text{Llama-3}_\text{8B}$~\citep{grattafiori2024llama3herdmodels}.
The average results for high- and low-resource languages are presented in Table~\ref{tab:different_scales}.
The results reveal that our \textit{ShifCon} framework continues to exhibit superior performance compared to MSFT. 
Specifically, in XGLM family models, \textit{ShifCon} demonstrates average improvements of 4.9\% and 4.5\% for the 564M and 2.9B scales, respectively. 
For BLOOM family models, \textit{ShifCon} shows average improvements of 4.1\% and 4.3\% for the 560M and 1.7B scales, respectively.
For $\text{Llama-3}_\text{8B}$, \textit{ShifCon} achieves an average improvement of 2.2\%, a relatively modest gain compared to other models. This can be attributed to the inherently stronger multilingual capabilities of $\text{Llama-3}_\text{8B}$. Nonetheless, the application of \textit{ShifCon} still brings benefits, particularly for low-resource languages. We believe this improvement is due to the notable performance gaps that remain for these languages, which our framework helps to mitigate.
Based on these observations, we derive the conclusion below: \textit{ShifCon} \emph{can generalize to models across different families and scales, which could be attributed to the selection of appropriate layers determined by the subspace distance metric.}

\input{tabs/language_error}

\begin{figure}[!t]
    \centering
    \centerline{\includegraphics[width=\columnwidth]{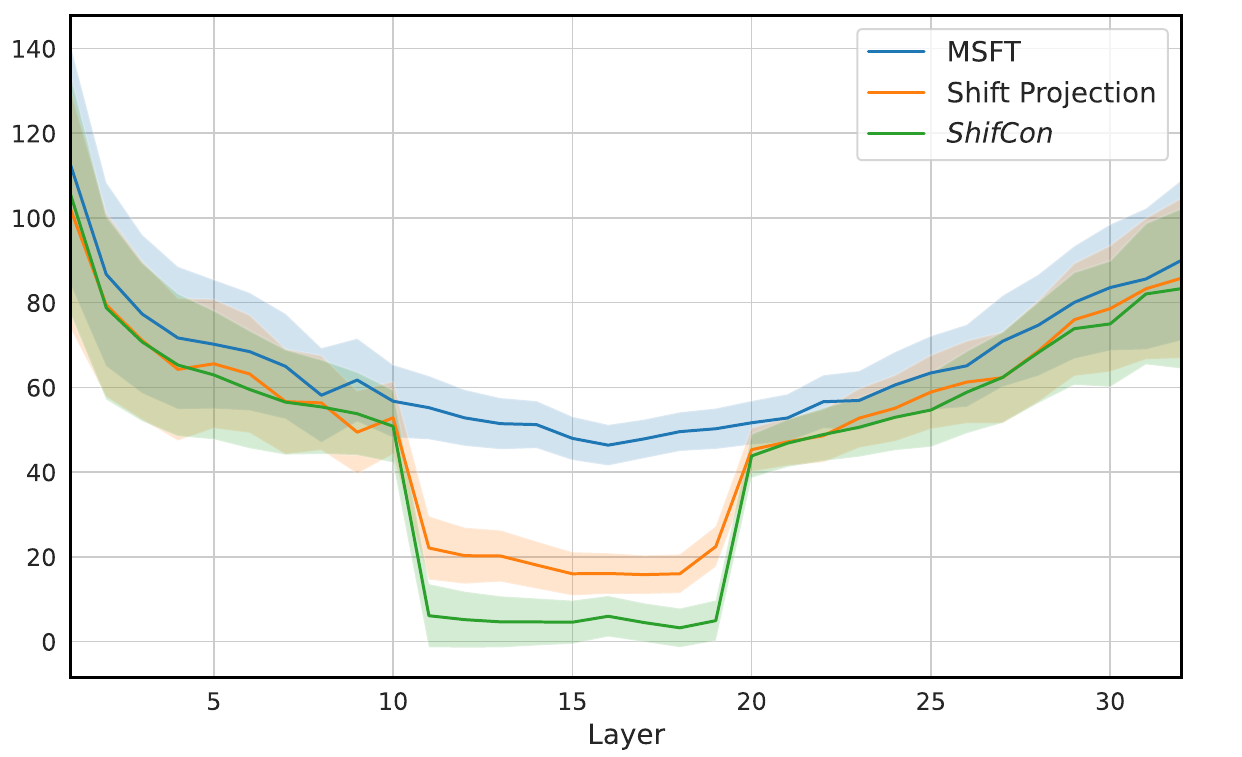}}
    \vspace{-0.2cm}
    \caption{The subspace distances of $\text{Llama-2}_\text{7B}$ after implementing shift projection and MCL.}
    \label{fig:exper-ablation}
    \vspace{-0.3cm}
\end{figure}

%% file: tabs/ablation_study.tex
\begin{table}[!t]
\renewcommand\arraystretch{1.1}

\centering
\setlength\tabcolsep{1.5pt}
\fontsize{9}{9}\selectfont 
\begin{tabular}{lccc}
\toprule[1.2pt]
                     & \multicolumn{1}{l}{$\text{Llama-2}_\text{7B}$} & \multicolumn{1}{l}{$\text{XGLM}_\text{7.5B}$} & \multicolumn{1}{l}{$\text{BLOOM}_\text{7.1B}$} \\ \addlinespace[2pt] \cline{1-4} \addlinespace[2pt]
\textit{ShifCon}                 & \textbf{46.0}                           & \textbf{43.8}                          & \textbf{44.1}                           \\
\ \ \ w/o Shift Projection & 42.2                           & 39.9                          & 40.7                           \\
\ \ \ w/o MCL              & 44.5                           & 43.1                          & 42.8                           \\ \bottomrule[1.2pt] 
\end{tabular}
\caption{The impact of Shift Projection and MCL in \textit{ShifCon} on the average results of all benchmarks. ``w/o'' means excluding this module from \textit{ShifCon}.}
\label{tab:ablation}
\end{table}

%% file: tabs/language_error.tex
\begin{table}[!t]
    \renewcommand\arraystretch{1.1}
    
    \centering
    \setlength\tabcolsep{1.5pt}
    \fontsize{9}{9}\selectfont 
    \begin{tabular}{lccc}
    \toprule[1.2pt]
                         & \multicolumn{1}{l}{$\text{Llama-2}_\text{7B}$} & \multicolumn{1}{l}{$\text{XGLM}_\text{7.5B}$} & \multicolumn{1}{l}{$\text{BLOOM}_\text{7.1B}$} \\ \addlinespace[2pt] \cline{1-4} \addlinespace[2pt]
    \textit{ShifCon}                 & \textbf{96.9}          & \textbf{97.6} & 94.9                           \\
    \ \ \ w/o Shift Projection & 87.6                           & 91.6                          & 88.8                                                 \\ 
    \ \ \ w/o MCL & 96.6                           & 97.3                          & \textbf{95.5}                                                 \\ 
    \bottomrule[1.2pt] 
    \end{tabular}
    \caption{The average results of the language consistency on the MGSM task. ``w/o'' means excluding this module from \textit{ShifCon}.}
    \label{tab:language_error}
    \end{table}

%% file: ablation_study/ablation_study.tex
Moreover, we investigate the impact of Shift Projection and MCL within \textit{ShifCon}. 
Table~\ref{tab:ablation} shows a performance decrease on ``\textit{ShifCon} w/o Shift Projection'', indicating that directly implementing MCL using original representations of non-dominant languages, instead of their dominant-like counterparts, leads to this decline.
We posit that applied MCL directly on original representations may compromise language-specific information within the representations, as it aims to bring representations of different languages with the same meaning closer together, making them become language-agnostic.

To explore this further, we follow~\citet{zhang-etal-2024-respond} to employ a language detector\footnote{\url{https://pypi.org/project/langdetect}} tool to assess the language consistency of input and output between \textit{ShifCon} and ``\textit{ShifCon} w/o Shift Projection''.
As shown in Table~\ref{tab:language_error}, a decrease in language consistency occurs when MCL is directly applied to the original representations.
Based on this observation, we give the following conclusion:

\begin{tcolorboxinsight}
    Directly applying MCL to original representations may compromise the language-specific information within representations, which impedes the model's ability to generate in that language, thereby adversely affecting performance.
\end{tcolorboxinsight}


Moreover, comparing \textit{ShifCon} and ``\textit{ShifCon} w/o MCL'', the performance increases. 
To delve deeper, we visualize the distribution of sentence representations and subspace distance between \textit{ShifCon} and ``\textit{ShifCon} w/o MCL'' in Fig.~\ref{fig:distribution_shift} and Fig.~\ref{fig:exper-ablation}, respectively.
The visualization reveals that:
\begin{tcolorboxinsight}
    MCL can further align the dominant-like representations of non-dominant language with its dominant language counterparts, thereby improving overall performance.   
\end{tcolorboxinsight}

%% file: low_subspace_area/low_subspace_area.tex
\begin{figure}[!t]
    \centering
    \centerline{\includegraphics[width=1.05\columnwidth]{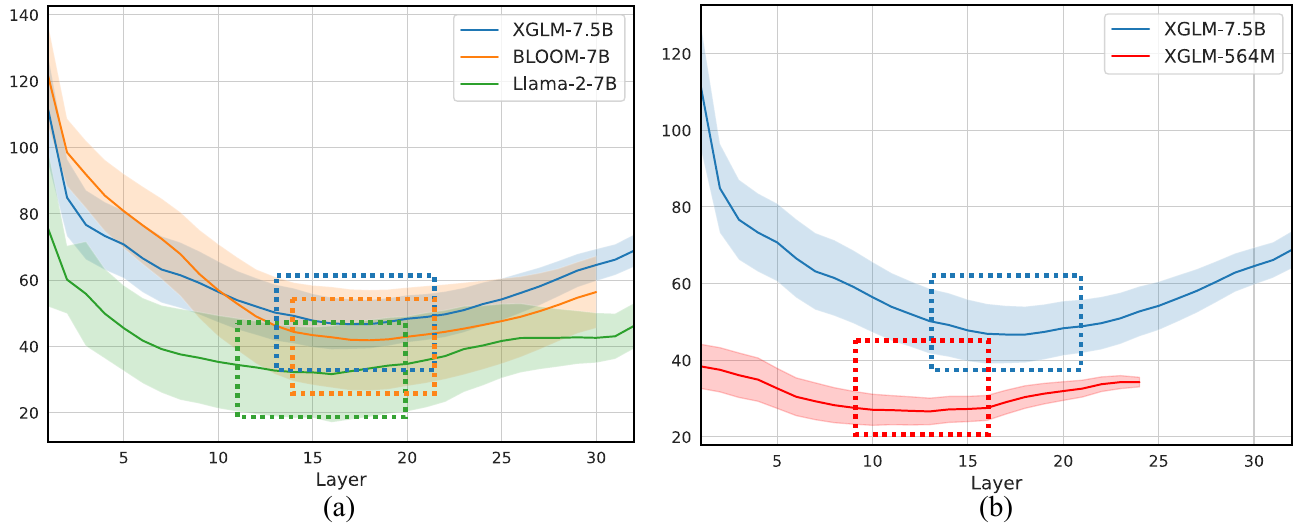}}
    \vspace{-0.2cm}
    \caption{The low subspace distance areas of different models are delineated with dashed boxes. (a) shows the results for different model families; (b) shows the results for different scales of XGLM. }
    \vspace{-0.3cm}
    \label{fig:exper-subspace-family}
\end{figure}
In Fig.~\ref{fig:exper-subspace-family}, we show the subspace distance areas of different models utilizing the $\beta$ value discovered in Finding~\ref{insight:2}. 
As depicted in Fig.~\ref{fig:exper-subspace-family} (a), we observe that the low subspace distance areas of $\text{Llama-2}_\text{7B}$, $\text{XGLM}_\text{7.5B}$, and $\text{BLOOM}_\text{7.1B}$ are [11, 20], [13, 22], and [14, 22] respectively.
This indicates that:
\begin{tcolorboxinsight}
    The low subspace distance areas of models from different families vary but generally locate in the middle and late-middle layers. 
\end{tcolorboxinsight}
Moreover, the subspace distances of $\text{XGLM}_\text{7.5B}$ and $\text{BLOOM}_\text{7.1B}$ are higher than $\text{Llama-2}_\text{7B}$, possibly due to they are being pre-trained on large-scale multilingual data, allowing them to learn more isolated representations for each language.

Another observation we find is that:
\begin{tcolorboxinsight}
    Models from the same family, despite having different layers, exhibit similar locations in the model for their low subspace distance areas. 
\end{tcolorboxinsight}

Specifically, in Fig.~\ref{fig:exper-subspace-family} (b), the low subspace distance areas of $\text{XGLM}_\text{7.5B}$ and $\text{XGLM}_\text{564M}$ are [13, 22] and [9, 16], respectively, both situated in the middle of the model.
Additionally, the subspace distance of $\text{XGLM}_\text{7.5B}$ is higher than $\text{XGLM}_\text{564M}$, possibly due to larger models showcasing enhanced language discrimination abilities.

%% file: lsd_effectiveness/lsd_effectiveness.tex
\begin{figure}[!t]
    \centering
    \centerline{\includegraphics[width=\columnwidth]{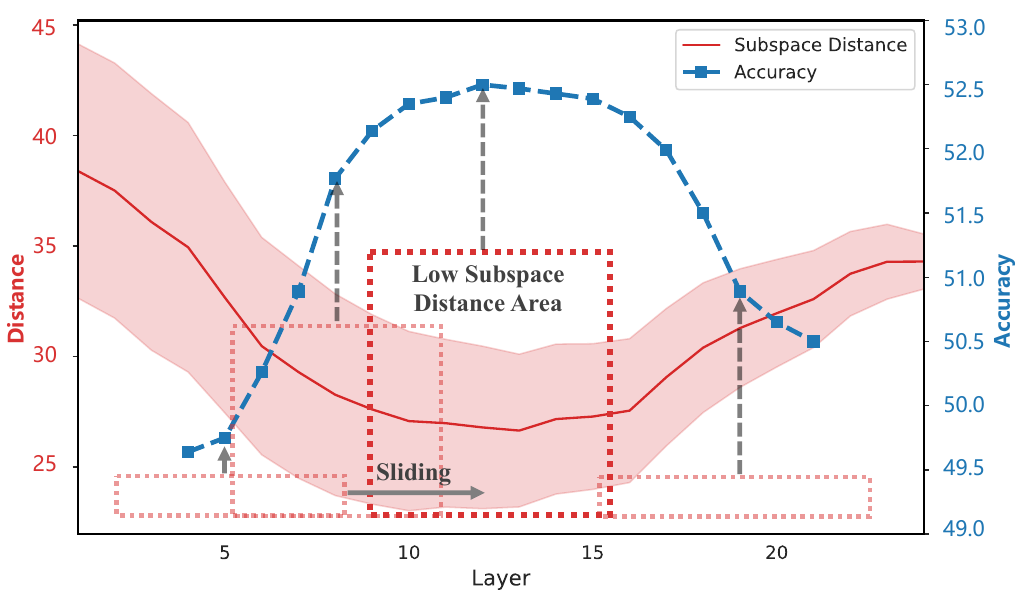}}
    \vspace{-0.2cm}
    \caption{The subspace distance of the $\text{XGLM}_\text{564M}$ and its average performance across three classification tasks using various layer areas. Each point's result denotes a model trained with the specific layer index as the medium of the layer area, such as the 5th layer index indicating a model trained with the [2, 8] layer area.}
    \label{fig:exper-acc-sub}
    \vspace{-0.2cm}
\end{figure}
We conduct extra experiments to verify if the layers within low subspace distance area are suitable for our \textit{ShifCon} framework.
Specifically, for the $\text{XGLM}_\text{564M}$ with 24 layers, we select $\lceil 24 \times 30\% \rceil = 8$ layers to apply our \textit{ShifCon}. 
We explore the performance of shift projection in regions beyond its low subspace distance area [9, 16] in a 8 layers sliding window manner.

As shown in Fig.~\ref{fig:exper-acc-sub}, as we slide the experimental layer area window from left to right, conducting \textit{ShifCon} in layer areas that exhibit great overlap with low subspace distance areas results in improved performance.
Moreover, as depicted in Fig.~\ref{fig:exper-subspace-family}, we find that the subspace distances of layers within the low subspace distance area are close.
This suggests that the language-specific information within the representations remains relatively unchanged, \emph{resulting in a stable distance between the subspaces of languages}. We speculate the model in these layers may focus on processing semantic information.
Based on these two observations, we give the following speculation:
\begin{tcolorboxinsight}
    Layers in the low subspace distance area are likely focused on information aggregation, thus aiding in gathering more information for non-dominant languages and enhancing performance.
\end{tcolorboxinsight}
This observation also highlights the effectiveness of our proposed distance metric (\S~\ref{sec:language_subspace}) in identifying the optimal layer area for our \textit{ShifCon}.

%% file: related/related.tex
\paragraph{Multilingual Bias in LLMs}
\label{LLM_bias}
Large Language Models (LLMs) have demonstrated remarkable multilingual capabilities as a result of their training on extensive and diverse multilingual datasets. These models have shown proficiency in various aspects of language processing across multiple languages, including multilingual reasoning, understanding, and generation~\citep{xue-etal-2021-mt5,lin-etal-2022-shot,anil2023palm2}.
However, empirical analysis indicates limited proficiency in low-resource languages, stemming from training data imbalances ~\citep{huang2023not,zhu2023multilingual,gurgurov2024multilingual}  and distinct representation spaces~\citep{wen-yi-mimno-2023-hyperpolyglot,liu-etal-2024-translico,yao2024data}.
Several studies have focused on scaling multilingual corpora through translation, which can provide preliminary capabilities for non-dominant languages. 
However, this approach is limited in both scale and quality due to the high cost of translated annotations and the presence of translation errors~\citep{muennighoff-etal-2023-crosslingual,zhang2023bayling,MathOctopus,tan2024large}.
In this study, we propose an internal alignment framework to further enhance the performance of non-dominant languages with limited MSFT data.

\paragraph{Representation Alignment}
\label{representation_alignment}
Previous studies have shown that projecting representations from the source to the target domain can mitigate domain discrepancies, facilitating effective cross-domain alignment and enhancing performance without disturbing the original domain subspace~\citep{kozhevnikov2014cross,chang-etal-2022-geometry,xu-etal-2023-language-representation,zhu2024eeg}.
However, this method often results in coarse alignment due to its unsupervised nature.
On the other hand, contrastive learning offers a more detailed representation learning approach by utilizing positive and negative pairs to encourage proximity within positive pairs and distance between negative pairs in a supervised manner. This method is better at capturing the complex relationships between representations and achieving precise alignment~\citep{radford2021learning,zhang2022fine,li2023multi,zhang2023assisting,zhang2024question,li-etal-2024-improving-context}.
Drawing from these insights, our framework first employs mean-shifted projection to map non-dominant language representations into the dominant language subspace, preserving language-specific information, and then applies contrastive learning for further alignment.

%% file: conclusion/conclusion.tex
This work aims to improve the performance of non-dominant languages with limited MSFT data.
To achieve this, we propose \textit{ShifCon} framework, which aims to align the internal forward process of non-dominant languages with that of the dominant language. 
It maps the representations of non-dominant languages into the dominant language's subspace to acquire their dominant-like representations, allowing them to access more information encoded in the model parameters.
The dominant-like representations are then shifted back to their native subspace to yield answers in their languages.
Furthermore, we propose a subspace distance metric to determine the optimal layer area for shift projection, and we apply multilingual contrastive learning to further enhance the internal alignment. 
The experimental results demonstrate that our proposed \textit{ShifCon} effectively improves the performance of non-dominant languages across models of various families and scales.
Our comprehensive analysis offers valuable insights for future research.

%% file: limitation.tex
The \textit{ShifCon} framework leverages translation pairs to conduct multilingual contrastive learning, which may pose challenges for low-resource languages or those lacking substantial parallel corpora.
Furthermore, due to computational resource limitations, the framework is restricted to multilingual generative language models with parameters not exceeding 8B.

Additionally, our forthcoming research endeavors will delve into exploring alternative model architectures, such as encoder-decoder models, to showcase the full potential and versatility of our proposed framework.

%% file: appendix.tex
\subsection{Visualization of Sentence Representations across Layers}
\label{sec:langs_subspace_across_layers}
\input{appendix/appendix_langs_subspace_across_layers}

\onecolumn\newpage
\subsection{Details of Language Subspace Distance}
\label{sec:detail_language_subspace_distance}

\input{appendix/appendix_language_subspace}

\subsection{Details of Evaluation}
\label{sec:datasets_and_evaluation}
\input{appendix/appendix_dataset_and_evaluation}

\onecolumn\newpage
\subsection{Implementation Details}
\label{sec:implementation_details}
\input{appendix/appendix_implementation_details.tex}

\onecolumn\newpage
\subsection{Detailed Results of Each Language across All the Benchmarks}
\label{sec:detailed_results}
\input{appendix/appendix_specific_results}

\onecolumn\newpage
\subsection{Low Subspace Distance Areas of Models across Different Families and Scales}
\label{sec:appendix_lsd}

\input{appendix/appendix_lsd_all}

\subsection{Language Code}
\label{sec:language_code}

\input{appendix/appendix_language_code.tex}

%% file: appendix/appendix_langs_subspace_across_layers.tex
\begin{figure}[!thbp]
    \centerline{\includegraphics[width=\columnwidth]{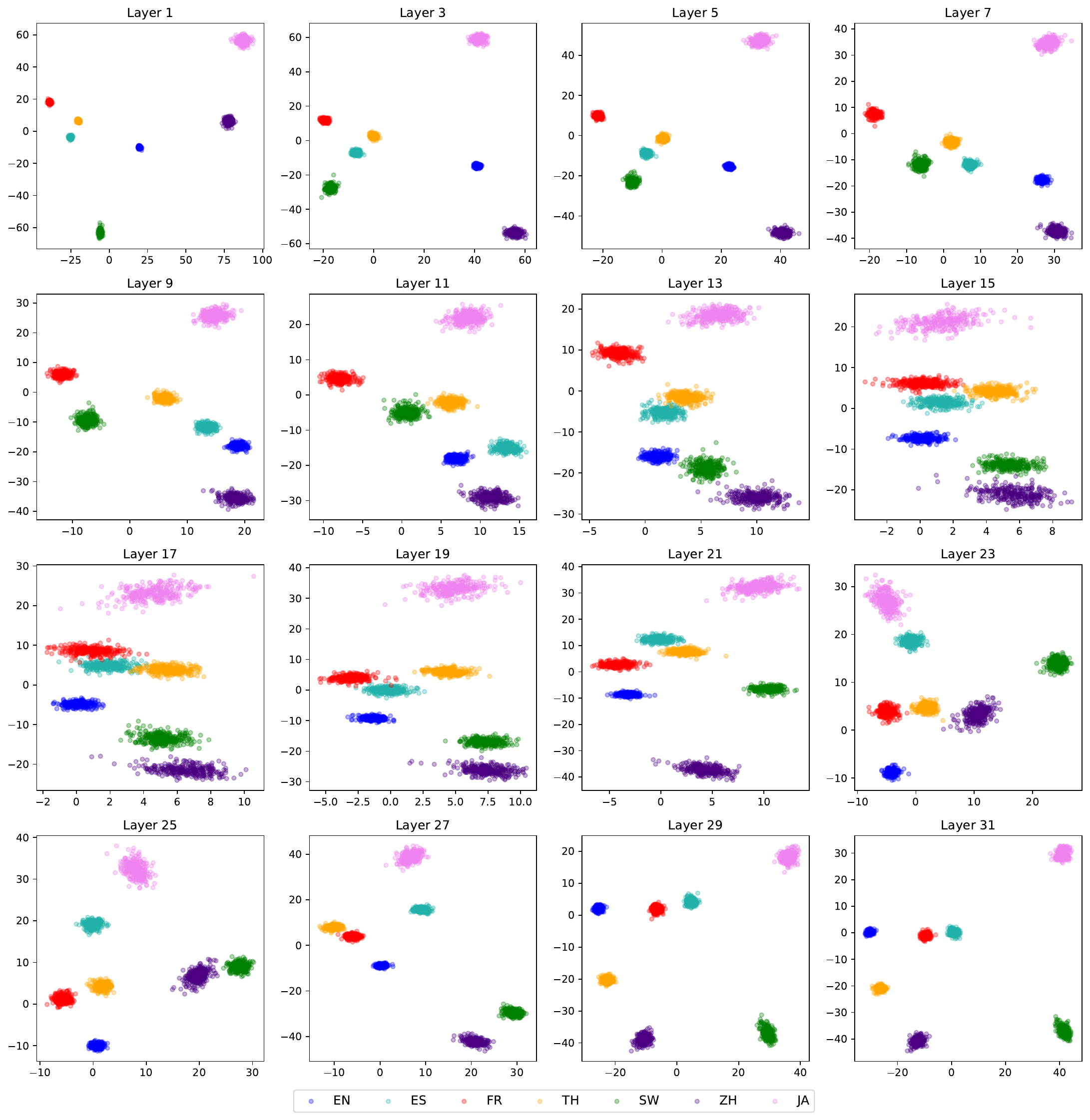}}
    \vspace{-0.2cm}
    \caption{We follow ~\citet{chang-etal-2022-geometry} to conduct LDA and present the visualization of sentence representations obtained by mean-pooling from $\text{Llama-2}_{7\text{B}}$ across layers along LDA components 1 and 3. We utilize 300 samples for each language from the FLORES dataset.}
    \label{fig:visualization_llama}
    \vspace{-0.3cm}
\end{figure}

\begin{figure}[!thbp]
    \centerline{\includegraphics[width=\columnwidth]{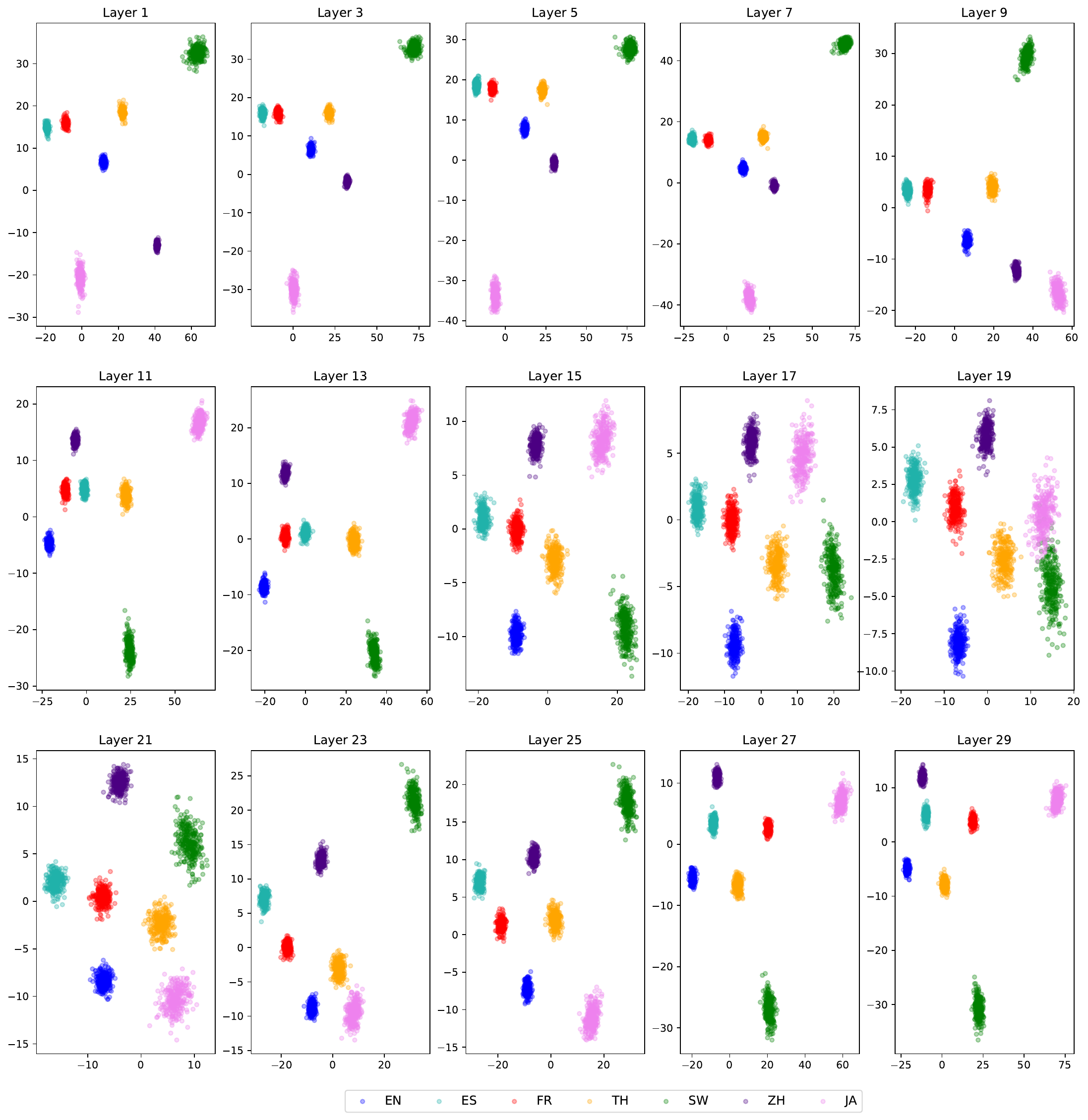}}
    \vspace{-0.2cm}
    \caption{We follow ~\citet{chang-etal-2022-geometry} to conduct LDA and present the visualization of sentence representations obtained by mean-pooling from $\text{BLOOM}_\text{7.1B}$ across layers along LDA components 1 and 3. We utilize 300 samples for each language from the FLORES dataset.}
    \label{fig:visualization_bloom}
    \vspace{-0.3cm}
\end{figure}

\begin{figure}[!thbp]
    \centerline{\includegraphics[width=\columnwidth]{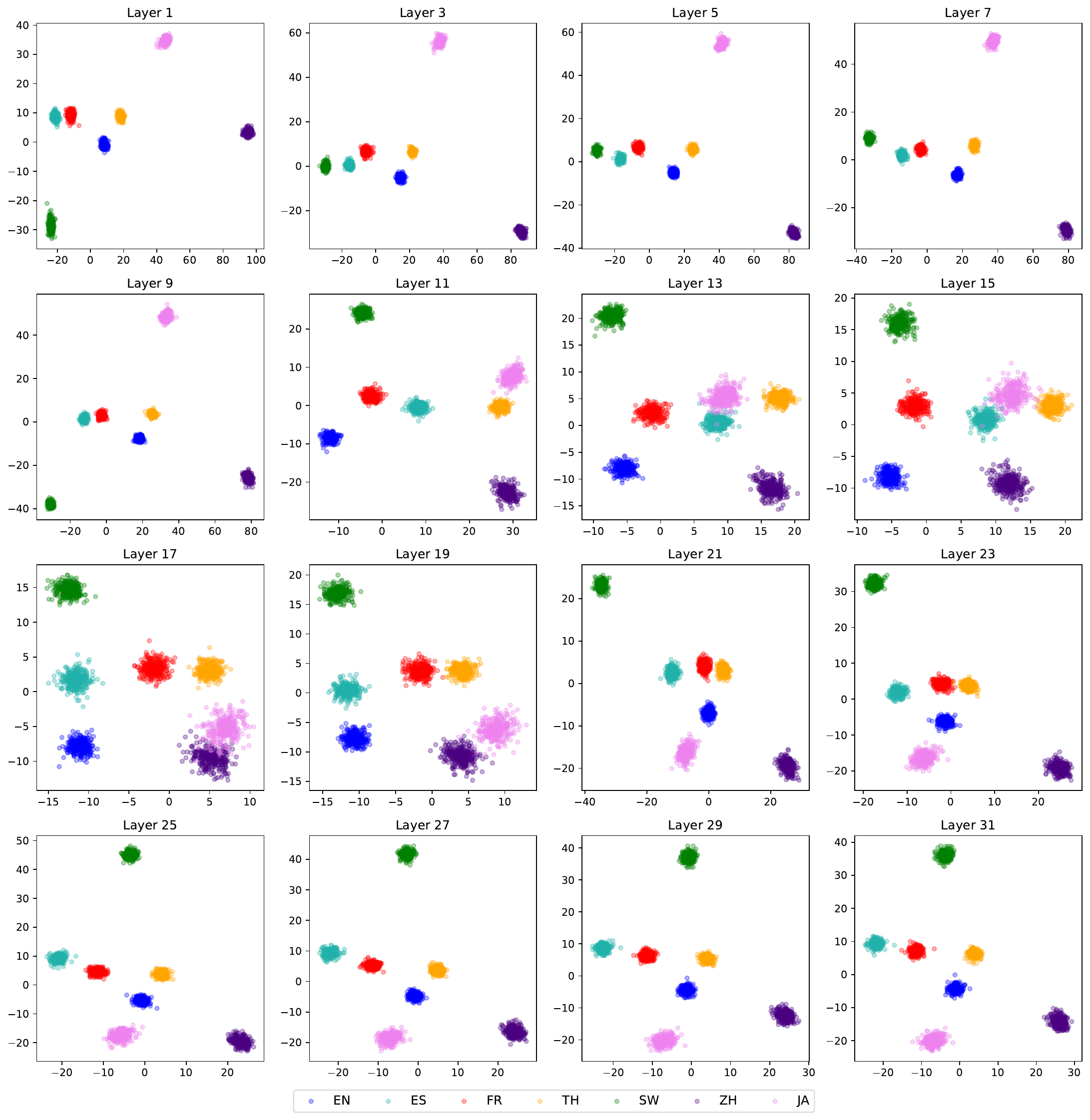}}
    \vspace{-0.2cm}
    \caption{We follow ~\citet{chang-etal-2022-geometry} to conduct LDA and present the visualization of sentence representations obtained by mean-pooling from $\text{XGLM}_\text{7.5B}$ across layers along LDA components 1 and 3. We utilize 300 samples for each language from the FLORES dataset.}
    \label{fig:visualization_xglm}
    \vspace{-0.3cm}
\end{figure}

%% file: appendix/appendix_language_subspace.tex
For each language $A$, we obtain a data matrix $\bm{X}_A \in \mathbb{R}^{n \times d}$ of $n$ contextualized token representations with $d$ dimensionality in language $A$ using 1k FLORES samples per language from the desired layer. 

The language subspace $\mathcal{S_A}$\footnote{We follow \citet{chang-etal-2022-geometry} to define the language subspace.} is described by the language's mean representation $\bm{\mu}_A \in \mathbb{R}^d$ along with $k$ principal directions of maximal variance in the language, defined by an orthonormal basis $\bm{V}_A \in \mathbb{R}^{d \times k_A}$.

In particular, $\bm{\mu}_A$ can be calculated as the mean value of $X_A$ along the token dimension $n$. 
As for $\bm{V}_A$, we first perform a singular value decomposition (SVD) of $\bm{X}_A$: $\bm{X}_A=\bm{U \Sigma V}^T$, where $\bm{U} \in \mathbb{R}^{n \times n}$ and $\bm{V} \in \mathbb{R}^{d \times d}$ are orthogonal. $\bm{\Sigma}\in \mathbb{R}^{n \times d}$ consists of a diagonal matrix $\bm{\Sigma}'\in \mathbb{R}^{d \times d}$ and a zero matrix, where $\bm{\Sigma}'=\text{diag}(\sigma_1, \sigma_2, \ldots, \sigma_d)$, with $\sigma_1 \geq \sigma_2 \geq \ldots  \geq \sigma_d \geq 0$. $\bm{\Sigma}'$ denotes the direction of greatest change in $\bm{X}_A$, which can be used for feature selecting. 
We select the first $k_A$ values to get $\bm{\Sigma}_A=\text{diag}(\sigma_1, \sigma_2, \ldots, \sigma_{k_A}) \in \mathbb{R}^{k_A \times k_A}$, while at the same time ensuring that the subspace accounted for 90\% of the total variance in the language.\footnote{Results were qualitatively similar for subspaces accounting for variance proportions in [75\%, 90\%, 95\%, 99\%].} 
Therefore, based on $\bm{\Sigma}_A$, we can obtain the corresponding $\bm{V}_A$ and leverage $\bm{U} \bm{\Sigma}_{A} \bm{V}_{A}^T$ to estimate $\bm{X}_A$.
Since $\bm{K}_A=\frac{1}{n-1}\bm{X}_{A}^{-1} \bm{X}_{A}$ \citep{chang-etal-2022-geometry}, the $\bm{K}_A\in \mathbb{R}^{d \times d}$ can be calculated with $\frac{1}{n-1} \bm{V}_{{A}} \bm{\Sigma}_{{A}}^2 \bm{V}_{{A}}^T$.

\subsection{Impact of Different Pooling Methods}
\label{sec:appendix_pooling_methods}
We also investigate the impact of three different pooling methods, namely mean-pooling, max-pooling, and last token representation, to derive sentence embeddings for our \emph{ShifCon} framework.

\begin{table}[!h]
\normalsize
\centering
\setlength \tabcolsep{4.5pt}
\begin{tabular}{lccc}
\toprule[1.2pt]
                    & $\text{Llama-2}_{\text{7B}}$ & $\text{XGLM}_{\text{7.5B}}$ & $\text{BLOOM}_{\text{7.1B}}$ \\ \hline
Mean-pooling                & 46.0       & 43.8      & 44.1        \\
Max-pooling & 45.2       & 43.3      & 43.6       \\
Last token & 45.8       & 44.1      & 43.7       \\ \bottomrule[1.2pt]
\end{tabular}
\caption{The average performance results of our $ShifCon$ framework across all benchmarks for the three different pooling methods.}
\label{tab:pooling_methods}
\end{table}

As demonstrated in Table~\ref{tab:pooling_methods}, the last token and mean pooling methods exhibit superior performance, and our approach shows less sensitivity to the choice of pooling method.

%% file: appendix/appendix_dataset_and_evaluation.tex
Due to the extensive training time required to train all languages included in Bactrian-X, we opt to sample a subset of representative languages, covering both high and low-resource languages for training. During evaluation, we focus on assessing the performance of the selected languages with corresponding benchmarks.
Detailed information regarding the languages used, evaluation metrics for each dataset are presented in Table~\ref{tab:benchmark}.
The evaluation prompt template are presented in Table~\ref{tab:template_evaluation}.

\definecolor{Color}{gray}{0.9}
\begingroup
\renewcommand{\arraystretch}{1.2} %
\begin{table*}[!htb]
\resizebox{\linewidth}{!}{
    \begin{tabular}{lllcc}
    \toprule
    \textbf{Dataset}      & \textbf{|Lang.|}          &\textbf{Languages}    & \textbf{Metric}& \textbf{Data Type}  \\ 
    \midrule
    Bactrian-X    &8   & English, Chinese, Indonesian, Spanish, Swahili, Thai, Turkish, Hindi & - & Train \\
    \rowcolor{Color}MGSM8KInstruct &10   & English, Chinese, Spanish, French, German, Russian, Japanese, Swahili, Thai, Bengali & - & Train \\
    MGSM &10   & English, Chinese, Spanish, French, German, Russian, Japanese, Swahili, Thai, Bengali & Accuracy & Test \\    
    \rowcolor{Color}XNLI      &7   & English, Spanish, Chinese, Turkish, Thai, Hindi, Swahili & Accuracy & Test \\
    XCOPA      &5   &  Chinese, Indonesian, Turkish, Thai, Swahili & Accuracy& Test\\
    \rowcolor{Color}XStoryCloze  &6   &English, Spanish, Chinese, Indonesian, Hindi, Swahili & Accuracy & Test \\
    FLORES  &6   &Spanish, Chinese, Indonesian, Turkish, Thai, Swahili & ChrF++ & Test \\
    
    \bottomrule
    \end{tabular}}
\label{tab:benchmark}
\caption{Multilingual datasets used in our experiments. We utilize ChrF++~\citep{popovic-2017-chrf} metric to evaluate the translation performance.}
\label{tab:benchmark}
\end{table*}
\endgroup

\definecolor{LightCyan}{rgb}{0.88,1,1}
\definecolor{TextColor}{rgb}{0.9,0.2,0.2}

\begingroup
\renewcommand{\arraystretch}{1.2} %

\begin{table*}[!h]
\centering
\resizebox{\linewidth}{!}{
    \begin{tabular}{lll}
    \toprule
    \textbf{Task}  & \textbf{Pattern}    & \textbf{Verbalizer}   \\ 
    \midrule
    XNLI
             & \texttt{\{premise\}} Based on the previous passage, is it true that 
             &  Yes || Maybe || No  \\
             &  \texttt{\{hypothesis\}}? Yes, No, or Maybe? \texttt{\{label\}} &  \\ 

    \rowcolor{LightCyan}XCOPA                  
            & \texttt{\{premise\}} \{\% if question == “cause" \%\}This happened because...&\\
    \rowcolor{LightCyan}
            &\{\% else \%\} As a consequence...\{\% endif \%\} &\\
    \rowcolor{LightCyan}
            &Help me pick the more plausible option:& \texttt{\{choice1\}} || \texttt{\{choice2\}} \\
    \rowcolor{LightCyan}
            & - \texttt{\{choice1\}}& \\
    \rowcolor{LightCyan}
            & - \texttt{\{choice2\}}& \\ 
    \rowcolor{LightCyan}
            & \texttt{\{label\}} & \\
    XStoryCloze   
            & \texttt{\{input\_sentence\_1\}} \texttt{\{input\_sentence\_2\}} 
            & \\
            &\texttt{\{input\_sentence\_3\}} \texttt{\{input\_sentence\_4\}}
            &\\
            & What is a possible continuation for the story given the following
            & \texttt{\{sentence\_quiz\_1\}} || \\
            & options?
            &\texttt{\{sentence\_quiz\_2\}} \\
            & - \texttt{\{sentence\_quiz\_1\}}
            & \\
           & - \texttt{\{sentence\_quiz\_2\}} &\\ 
           &\texttt{\{label\}} 
           & \\
    \rowcolor{LightCyan}
    FLORES
          & Translate the following \texttt{\{src\_language\}} text to \texttt{\{tgt\_language\}}: & \texttt{\{tgt\_sentence\}}\\
          \rowcolor{LightCyan}
          & \texttt{\{src\_sentence\}} \texttt{\{tgt\_sentence\}} & \\
    \bottomrule
    \end{tabular}
}
\caption{The prompt templates used for evaluation following \citet{muennighoff-etal-2023-crosslingual} and \citet{zhang-etal-2024-impact}.}
\label{tab:template_evaluation}
\end{table*}

%% file: appendix/appendix_implementation_details.tex
\paragraph{Model Information}

\begin{table*}[!h]
\normalsize
\centering
\setlength \tabcolsep{4.5pt}
\begin{tabular}{lccc}
\toprule[1.2pt]
           & \textbf{Dimension} & \textbf{Heads} & \textbf{Layers} \\ \cline{2-4} 
$\text{Llama-2}_{7\text{B}}$ & 4096               & 32             & 32              \\ 
$\text{Llama-3}_{8\text{B}}$ & 4096               & 32             & 32              \\ \addlinespace[2pt] \hdashline[1pt/1pt] \addlinespace[2pt]
$\text{BLOOM}_{\text{7.1B}}$ & 4096               & 32             & 30  \\    
$\text{BLOOM}_{\text{1.7B}}$ & 2048               & 16             & 24              \\
$\text{BLOOM}_{\text{560M}}$ & 1024               & 16             & 24              \\  \addlinespace[2pt] \hdashline[1pt/1pt] \addlinespace[2pt]
$\text{XGLM}_{7.5\text{B}}$  & 4096               & 32             & 32              \\
$\text{XGLM}_{2.9\text{B}}$  & 2048               & 16             & 48              \\
$\text{XGLM}_{564\text{M}}$  & 1024               & 16             & 24              \\ \bottomrule[1.2pt]
\end{tabular}
\vspace{-0.1cm}
\caption{The detailed information of the models utilized in our experiment. ``Dimension'', ``Heads'', and ``Layers'' denote the dimension of representation, attention heads, and number of layers, respectively.}
\label{tab:model_info}
\vspace{-0.2cm}
\end{table*}

In Table~\ref{tab:model_info}, we provide comprehensive details about the models utilized in our experiment. 
Here, ``Dimension'', ``Heads'', and ``Layers'' represent the representation dimension, attention heads, and number of layers, respectively.

\paragraph{Training Settings}
Our experiments are conducted with 4xA100 GPUs. 
Each experiment is run with three different random seeds, and the results are averaged to obtain the final outcome.
The temperature $\tau$ is set to 0.05 in the multilingual contrastive learning procedure. 
We follow previous multitasking works~\citep{kong2022blcu,kong2022multitasking,zhang2023assisting,zhang2025guilomo} to explore $\alpha$ values in Eq.~\ref{eq:total_loss} within [0.5, 1.0, 1.5, 2.0] to determine the best performance.
Following the training settings from previous works~\citep{li2024contextualization,li2024dalk,tong2024can,wang2024bpo}, we set the learning rate for training models with parameters exceeding 7 billion to 1e-5, while for others to 3e-5. We set the maximum sequence length to 512 and the global batch size to 128. 
In generation tasks, we utilize a greedy decoding strategy to help replicate our results accurately. 
A cosine scheduler with a 3\% warm-up period is implemented.
Mixed precision training and ZeRO are employed within the DeepSpeed training framework to accelerate the training process and conserve memory usage. 
The AdamW~\citep{loshchilov2018decoupled} optimizer is utilized to update the model parameters during the training process.

For the AFP baseline method, we adhere to the training configuration outlined by~\citet{li-etal-2024-improving-context} to train the models. Specifically, we define $p_{src}$ for cross-lingual guidance during training and perform multilingual contrastive learning on the first layer.

Additionally, we explore our \textit{ShifCon} framework with a two-stage training strategy, which involves initial training solely with MSFT loss to establish a preliminary model, followed by further fine-tuning using our \textit{shifCon} framework.
As depicted in Table~\ref{tab:two_stage_results}, the results indicate that implementing a two-stage training strategy leads to better performance.
We posit that the preliminary model obtained by MSFT in the first stage could offer better representations for each language, facilitating shift projection and multilingual contrastive learning.
Consequently, all results are reported based on the two-stage training strategy in our paper.

\begin{table}[!h]
\normalsize
\centering
\setlength \tabcolsep{4.5pt}
\begin{tabular}{lccc}
\toprule[1.2pt]
                    & $\text{Llama-2}_{\text{7B}}$ & $\text{XGLM}_{\text{7.5B}}$ & $\text{BLOOM}_{\text{7.1B}}$ \\ \hline
MSFT                & 43.8       & 41.6      & 42.2       \\
\textit{ShifCon} w/ Two-Stage & 46.0       & 43.8      & 44.1       \\
\textit{ShifCon} w/ One-Stage & 44.8       & 41.7      & 42.5       \\ 
\bottomrule[1.2pt]
\end{tabular}
\caption{The average performance results of our $ShifCon$ framework across all benchmarks for the three model families, comparing the two-stage and one-stage training strategies.}
\label{tab:two_stage_results}
\end{table}

\subsection{New Strategy for Obtaining Better Language Vectors}
\label{sec:appendix_new_strategy}
Given that model parameters are updated at each training step, it is essential for the language vectors to be updated correspondingly. Inspired by the batch normalization paradigm, we introduce a novel strategy aimed at improving the quality of language vectors.
As calculating the mean representation of all samples in language $a$ after updating parameters for each batch is computationally expensive, we utilize the mean representation of language $a$ samples in the $t$-th batch to estimate.
Specifically, for the representations of language $a$ in $t$-th batch at $l$-th layer, let $\bm{v}_t$ denote the mean representation of language $a$ samples from first batch to $t$-th batch and $\bm{u}_t$ denote the mean representation of the samples in language $a$ from the $t$-th batch (Noted that, $\bm{v}_t$ is computed by $t$-th step's model). 
The estimation of $\bm{v}_t$, i.e., $\hat{\bm{v}}_t$, can be obtained by using the representations of $t$-th batch computed by corresponding $t$-th step's model:
\begin{equation}
    \hat{\bm{v}}_t = \frac{\sum_{i=1}^t \eta^{i-1} \bm{u}_i}{\sum_{i=1}^t \eta^{i-1}}
\end{equation}

\noindent where $\eta \geq 1$ denotes the enhancement factor. $\eta^{i-1}$ denotes the $i-1$-th power of $\eta$.
As $t$ increases, the model becomes more accurate, leading to more precise representation $\bm{u}_t$. Consequently, the corresponding weight factors are larger.

Subsequently, we can estimate the mean representation of next batch's  $\bm{v}_t$ through the following approach:
\begin{equation}
\begin{split}
    \hat{\bm{v}}_{t+1} &= \frac{\sum_{i=1}^{t+1} \eta^{i-1} \bm{u}_i}{\sum_{i=1}^{t+1} \eta^{i-1}} \\
    &= \frac{1}{\sum_{i=1}^{t+1} \eta^{i-1}} \eta^t \bm{u}_{t+1}
    + \frac{\sum_{i=1}^{t} \eta^{i-1}}{\sum_{i=1}^{t+1} \eta^{i-1}} \Big( \frac{1}{\sum_{i=1}^{t} \eta^{i-1}} \sum_{i=1}^{t} \eta^{i-1} \Big) \\
    &= \frac{\eta^t}{\sum_{i=0}^{t} \eta^{i}} \bm{u}_{t+1}
    + \frac{\sum_{i=0}^{t-1} \eta^{i}}{\sum_{i=0}^{t} \eta^{i}} \hat{\bm{v}}_{t}
\end{split}
\end{equation}
Here, we only need the estimated mean representation $\hat{\bm{v}}_{t}$ and the true mean representation of the samples from the $t+1$ batch $\bm{u}_{t+1}$, to generate an estimation of the mean representation of $\hat{\bm{v}}_{t+1}$.
For simplicity, we directly set $\frac{\eta^t}{\sum_{i=0}^{t} \eta^{i}} = \frac{1}{4}$ and $\frac{\sum_{i=0}^{t-1} \eta^{i}}{\sum_{i=0}^{t} \eta^{i}} = \frac{3}{4}$ in this work.

We conduct an extra ablation experiment on $\text{XGLM}_\text{564M}$ to verify the effectiveness of our proposed strategy. As the experimental results shown in Table~\ref{tab:appendix_new_strategy}, when compared with the straightforward method, that is, simply mean pooling the representations, our strategy can yield better performance.
\input{tabs/appendix_new_strategy.tex}

%% file: tabs/appendix_new_strategy.tex
\begin{table}[!h]
    \renewcommand\arraystretch{1.1}
    \centering
    
    \setlength\tabcolsep{4pt}
    \fontsize{10}{12}\selectfont 
    
    \begin{tabular}{lcccccccc}
    \toprule[1.2pt]
                      & \multicolumn{2}{c}{\textbf{XCOPA}} & \textbf{} & \multicolumn{2}{c}{\textbf{XNLI}} & \textbf{} & \multicolumn{2}{c}{\textbf{XStoryCloze}} \\  \cline{2-3} \cline{5-6} \cline{8-9} \addlinespace[2pt]
                      & \textbf{High}    & \textbf{Low}    & \textbf{} & \textbf{High}   & \textbf{Low}    & \textbf{} & \textbf{High}       & \textbf{Low}       \\ \addlinespace[2pt] \hline \addlinespace[2pt] 
    w/ New Strategy         & 58.4             & 55.8            &           & 42.6            & 40.5            &           & 59.8                & 58.1               \\
    w/ Mean Pooling              & 58.1             & 55.3            &           & 42.3            & 40.1            &           & 59.6                & 57.6                 \\
     \bottomrule[1.2pt] 
    \end{tabular}
    \caption{The average performance of high- and low-resource languages across three classification tasks with two different language vector strategies.}
    \label{tab:appendix_new_strategy}
    \vspace{-0.2cm}
    \end{table}

%% file: appendix/appendix_specific_results.tex
\input{tabs/appendix_mgsm}

\input{tabs/appendix_flores}
\input{tabs/appendix_xcopa}

\input{tabs/appendix_xnli}

\input{tabs/appendix_xstorycloze}

%% file: tabs/appendix_mgsm.tex
\begin{table*}[!th]
    \renewcommand\arraystretch{1.1}
    
    \centering
    
    \setlength\tabcolsep{4pt}
    \fontsize{10}{12}\selectfont 
    
    \begin{tabular}{lccccccclccc}
    \toprule[1.2pt]
                     & \multicolumn{7}{c}{High}                       &  & \multicolumn{3}{c}{Low} \\ \addlinespace[2pt] \cline{2-8} \cline{10-12} \addlinespace[2pt]
                     & EN   & ZH   & DE   & ES   & FR   & JA   & RU   &  & SW     & BN     & TH    \\ \hline \addlinespace[2pt]
    $\text{Llama-2}_\text{7B}$   & 51.4 & 29.6 & 37.2 & 34.8 & 36.4 & 26.2 & 30.8 &  & 2.8    & 7.2    & 5.2   \\
     \ \ \ \ \ \ \ \ \ +MSFT             & 59.8 & 43.2 & 45.2 & 46.0   & 42.4 & 34.4 & 43.6 &  & 31.6   & 22.8   & 34.2  \\
      \ \ \ \ \ \ \ \ \ +AFP             & 60.0 & 42.8 & 46.4 & 47.2   & 45.6 & 37.2 & 45.2 &  & 34.4   & 25.2   & 35.6  \\
     \ \ \ \ \ \ \ \ \ +\textit{ShifCon} & 58.2 & 48.4 & 48.8 & 45.6 & 47.2 & 40.4 & 48.8 &  & 38.0   & 28.4   & 38.8  \\  \addlinespace[2pt] \hdashline[1pt/1pt] \addlinespace[2pt]
    $\text{XGLM}_\text{7.5B}$   & 7.6  & 4.8  & 3.6  & 3.2  & 2.8  & 2.8  & 2.8  &  & 1.2    & 2.0    & 2.4   \\
     \ \ \ \ \ \ \ \ \ +MSFT             & 14.4 & 9.6  & 10.0 & 10.4 & 10.8 & 8.0  & 10.8 &  & 6.8    & 7.2    & 6.8   \\
      \ \ \ \ \ \ \ \ \ +AFP             & 16.4 & 9.2  & 12.4 & 13.2 & 12.4 & 11.2  & 10.4 &  & 9.6    & 9.2    & 10.0   \\
     \ \ \ \ \ \ \ \ \ +\textit{ShifCon} & 15.6 & 12.8 & 14.0 & 12.8 & 15.2 & 12.0 & 13.6 &  & 11.2   & 11.6   & 14.4  \\ \bottomrule[1.2pt]  
    \end{tabular}
    \caption{The detailed results of each language on the MGSM task in $\text{Llama-2}_\text{7B}$ and $\text{XGLM}_\text{7.5B}$. High- and low-resource languages are categorized based on their data ratios in the pre-training corpus.}
    \end{table*}

    \begin{table*}[!th]
    \renewcommand\arraystretch{1.1}
    
    \centering
    
    \setlength\tabcolsep{4pt}
    \fontsize{10}{12}\selectfont 
    \begin{tabular}{lccccccclccc}
    \toprule[1.2pt]
                     & \multicolumn{4}{c}{High}                       &  & \multicolumn{6}{c}{Low} \\ \addlinespace[2pt] \cline{2-5} \cline{7-12} \addlinespace[2pt]
                     & EN   & ZH   & ES   & FR  & & SW   & BN   & TH    & DE     & JA     & RU    \\ \hline \addlinespace[2pt]
    $\text{BLOOM}_\text{7.1B}$   & 20.0 & 9.2 & 11.6 & 12.0 & & 2.4 & 5.2 & 1.6  & 4.0    & 2.4    & 6.8   \\
     \ \ \ \ \ \ \ \ \ +MSFT             & 26.8 & 18.8 & 21.6 & 20.4 &  & 11.6 & 13.2 & 10.4  & 13.6   & 12.4   & 14.0  \\
      \ \ \ \ \ \ \ \ \ +AFP             & 28.4 & 18.0 & 23.2 & 22.0 &  & 14.8 & 15.6 & 14.4  & 15.2   & 16.4   & 18.0  \\
     \ \ \ \ \ \ \ \ \ +\textit{ShifCon} & 28.0 & 21.2 & 24.8 & 24.0 &  & 19.2 & 18.8 & 17.6  & 19.6   & 18.4   & 19.4   \\  \bottomrule[1.2pt]  
    \end{tabular}
    \caption{The detailed results of each language on the MGSM task in $\text{BLOOM}_\text{7.1B}$. High- and low-resource languages are categorized based on their data ratios in the pre-training corpus.}
    \end{table*}

%% file: tabs/appendix_flores.tex
\begin{table*}[!th]
    \renewcommand\arraystretch{1.1}
    
    \centering
    
    \setlength\tabcolsep{4pt}
    \fontsize{10}{12}\selectfont 
    \begin{tabular}{lccccccc}
    \toprule[1.2pt]
                     & \multicolumn{3}{c}{High} &  & \multicolumn{3}{c}{Low} \\ \addlinespace[2pt]\cline{2-4} \cline{6-8} \addlinespace[2pt]
                     & ES     & ZH     & ID     &  & SW     & TH     & TR    \\ \hline \addlinespace[2pt]
    $\text{Llama-2}_\text{7B}$ & 42.6   & 17.1   & 40.9   &  & 14.7   & 12.9   & 20.0  \\
     \ \ \ \ \ \ \ \ \ +MSFT             & 43.4   & 18.9   & 41.8   &  & 18.1   & 15.4   & 21.8  \\
     \ \ \ \ \ \ \ \ \ +AFP              & 43.9   & 19.5   & 42.4   &  & 18.9   & 16.2   & 22.4  \\
     \ \ \ \ \ \ \ \ \ +\textit{ShifCon} & 44.5   & 19.8   & 42.6   &  & 20.2   & 16.6   & 22.3  \\ \addlinespace[2pt] \hdashline[1pt/1pt] \addlinespace[2pt]
    $\text{XGLM}_\text{7.5B}$      & 36.1   & 17.8   & 42.9   &  & 33.2   & 31.5   & 30.0  \\
     \ \ \ \ \ \ \ \ \ +MSFT             & 36.8   & 19.1   & 44.5   &  & 35.7   & 32.1   & 30.5  \\
     \ \ \ \ \ \ \ \ \ +AFP              & 37.4   & 19.8   & 45.0   &  & 35.9   & 32.9   & 31.2  \\
     \ \ \ \ \ \ \ \ \ +\textit{ShifCon} & 37.8   & 20.8   & 44.9   &  & 36.8   & 33.8   & 31.8  \\ \addlinespace[2pt] \hdashline[1pt/1pt] \addlinespace[2pt]
    $\text{BLOOM}_\text{7.1B}$      & 40.2   & 35.2   & 48.8   &  & 37.1   & 16.2   & 19.6  \\
     \ \ \ \ \ \ \ \ \ +MSFT             & 40.5   & 36.0   & 50.5   &  & 37.8   & 17.6   & 22.3  \\
     \ \ \ \ \ \ \ \ \ +AFP              & 41.2   & 36.9   & 51.1   &  & 38.5   & 18.2   & 23.1  \\
     \ \ \ \ \ \ \ \ \ +\textit{ShifCon} & 41.6   & 36.8   & 51.7   &  & 39.2   & 18.4   & 23.9  \\ \bottomrule[1.2pt]  
    \end{tabular}
    \caption{The detailed results of each language on the FLORES (en-xx) task in $\text{Llama-2}_\text{7B}$, $\text{XGLM}_\text{7.5B}$, and $\text{BLOOM}_\text{7.1B}$. 
    High- and low-resource languages are categorized based on their data ratios in the pre-training corpus.}
    \end{table*}

    \begin{table*}[!th]
    \renewcommand\arraystretch{1.1}
    
    \centering
    
    \setlength\tabcolsep{4pt}
    \fontsize{10}{12}\selectfont 
    \begin{tabular}{lccccccc}
    \toprule[1.2pt]
                      & \multicolumn{3}{c}{High} &  & \multicolumn{3}{c}{Low} \\ \addlinespace[2pt] \cline{2-4} \cline{6-8} \addlinespace[2pt]
                     & ES     & ZH     & ID     &  & SW     & TH     & TR    \\ \hline \addlinespace[2pt]
    $\text{Llama-2}_\text{7B}$       & 49.2   & 18.8   & 51.5   &  & 23.5   & 11.6   & 29.1  \\
     \ \ \ \ \ \ \ \ \ +MSFT             & 48.6   & 19.4   & 53.2   &  & 26.9   & 16.4   & 30.8  \\
     \ \ \ \ \ \ \ \ \ +AFP              & 49.0   & 19.7   & 54.4   &  & 27.5   & 17.1   & 31.6  \\
     \ \ \ \ \ \ \ \ \ +\textit{ShifCon} & 49.5   & 21.2   & 54.8   &  & 29.4   & 17.8   & 32.5  \\ \addlinespace[2pt] \hdashline[1pt/1pt] \addlinespace[2pt]
     $\text{XGLM}_\text{7.5B}$      & 41.8   & 33.4   & 48.3   &  & 42.9   & 26.7   & 37.9  \\
     \ \ \ \ \ \ \ \ \ +MSFT             & 43.0   & 33.9   & 50.1   &  & 43.9   & 28.3   & 39.6  \\
     \ \ \ \ \ \ \ \ \ +AFP              & 44.0   & 34.8   & 51.2   &  & 44.5   & 28.9   & 39.9  \\
     \ \ \ \ \ \ \ \ \ +\textit{ShifCon} & 43.8   & 35.6   & 51.7   &  & 45.2   & 29.8   & 40.4  \\ \addlinespace[2pt] \hdashline[1pt/1pt] \addlinespace[2pt]
    $\text{BLOOM}_\text{7.1B}$    & 45.8   & 39.6   & 51.6   &  & 43.8   & 20.3   & 28.1  \\
     \ \ \ \ \ \ \ \ \ +MSFT             & 46.4   & 39.9   & 53.3   &  & 45.4   & 23.0   & 30.8  \\
     \ \ \ \ \ \ \ \ \ +AFP              & 46.9   & 40.5   & 53.8   &  & 46.0   & 23.7   & 31.6  \\
     \ \ \ \ \ \ \ \ \ +\textit{ShifCon} & 47.6   & 41.3   & 52.8   &  & 46.8   & 24.5   & 32.4  \\ \bottomrule[1.2pt]  
    \end{tabular}
    \caption{The detailed results of each language on the FLORES (xx-en) task in $\text{Llama-2}_\text{7B}$, $\text{XGLM}_\text{7.5B}$, and $\text{BLOOM}_\text{7.1B}$. 
    High- and low-resource languages are categorized based on their data ratios in the pre-training corpus.}
    \end{table*}

%% file: tabs/appendix_xcopa.tex
\begin{table*}[!th]
    \renewcommand\arraystretch{1.1}
    
    \centering
    
    \setlength\tabcolsep{4pt}
    \fontsize{10}{12}\selectfont 
    \begin{tabular}{lcccccc}
    \hline
                     & \multicolumn{2}{c}{High} & \multicolumn{1}{l}{} & \multicolumn{3}{c}{Low} \\ \addlinespace[2pt] \cline{2-3} \cline{5-7} \addlinespace[2pt]
                     & ZH          & ID         & \multicolumn{1}{l}{} & TR     & TH     & SW    \\ \hline \addlinespace[2pt]
    $\text{Llama-2}_\text{7B}$      & 63.8        & 62.6       &                      & 49.0   & 51.4   & 48.8  \\
    \ \ \ \ \ \ \ \ \ +MSFT             & 65.0        & 63.4       &                      & 51.8   & 52.6   & 51.5  \\
    \ \ \ \ \ \ \ \ \ +AFP              & 65.8        & 64.2       &                      & 52.9   & 53.4   & 52.3  \\
    \ \ \ \ \ \ \ \ \ +\textit{ShifCon} & 66.8        & 64.2       &                      & 54.1   & 53.2   & 53.2  \\ \addlinespace[2pt] \hdashline[1pt/1pt] \addlinespace[2pt]
    $\text{XGLM}_\text{7.5B}$      & 63.6        & 64.0       &                      & 56.8   & 57.1   & 58.2  \\
    \ \ \ \ \ \ \ \ \ +MSFT             & 64.4        & 65.4       &                      & 58.4   & 58.8   & 57.6  \\
    \ \ \ \ \ \ \ \ \ +AFP              & 65.3        & 66.2       &                      & 59.3   & 59.2   & 58.3  \\
    \ \ \ \ \ \ \ \ \ +\textit{ShifCon} & 66.8        & 66.8       &                      & 60.2   & 59.4   & 60.6  \\ \addlinespace[2pt] \hdashline[1pt/1pt] \addlinespace[2pt]
    $\text{BLOOM}_\text{7.1B}$     & 57.1        & 58.4       &                      & 53.2   & 50.8   & 52.1  \\
    \ \ \ \ \ \ \ \ \ +MSFT             & 58.6        & 59.8       &                      & 55.5   & 51.6   & 54.6  \\
    \ \ \ \ \ \ \ \ \ +AFP              & 59.4        & 60.5       &                      & 56.3   & 52.8   & 55.5  \\
    \ \ \ \ \ \ \ \ \ +\textit{ShifCon} & 60.2        & 60.4       &                      & 57.6   & 54.4   & 56.8  \\ \bottomrule[1.2pt]
    \end{tabular}
    \caption{The detailed results of each language on the XCOPA task in $\text{Llama-2}_\text{7B}$, $\text{XGLM}_\text{7.5B}$, and $\text{BLOOM}_\text{7.1B}$. 
    High- and low-resource languages are categorized based on their data ratios in the pre-training corpus.}
    \end{table*}

%% file: tabs/appendix_xnli.tex
\begin{table*}[!th]
    \renewcommand\arraystretch{1.1}
    
    \centering
    \setlength\tabcolsep{4pt}
    \fontsize{10}{12}\selectfont 
    \begin{tabular}{lccclcccc}
    \toprule[1.2pt]
                     & \multicolumn{3}{c}{High} &  & \multicolumn{4}{c}{Low}   \\ \addlinespace[2pt]\cline{2-4} \cline{6-9} \addlinespace[2pt]
                     & EN     & ES     & ZH     &  & TR   & TH   & HI   & SW   \\ \hline \addlinespace[2pt]
    $\text{Llama-2}_\text{7B}$       & 49.1   & 42.6   & 44.0   &  & 35.8 & 37.2 & 37.1 & 30.8 \\
     \ \ \ \ \ \ \ \ \ +MSFT             & 50.8   & 43.8   & 44.5   &  & 37.5 & 39.5 & 38.8 & 34.6 \\
      \ \ \ \ \ \ \ \ \ +AFP             & 50.8   & 44.4   & 45.3   &  & 38.6 & 40.7 & 39.6 & 35.8 \\
     \ \ \ \ \ \ \ \ \ +\textit{ShifCon} & 50.4   & 45.0   & 46.1   &  & 40.8 & 41.8 & 40.2 & 38.1 \\ \addlinespace[2pt] \hdashline[1pt/1pt] \addlinespace[2pt]
    $\text{XGLM}_\text{7.5B}$      & 46.9   & 41.6   & 45.0   &  & 39.8 & 43.2 & 42.6 & 40.1 \\
     \ \ \ \ \ \ \ \ \ +MSFT             & 48.7   & 42.4   & 46.7   &  & 41.3 & 44.4 & 42.2 & 41.2 \\
      \ \ \ \ \ \ \ \ \ +AFP             & 49.9   & 43.3   & 47.8   &  & 43.1 & 45.2 & 43.1 & 42.0 \\
     \ \ \ \ \ \ \ \ \ +\textit{ShifCon} & 51.2   & 45.8   & 48.9   &  & 44.7 & 44.8 & 43.8 & 43.8 \\ \addlinespace[2pt] \hdashline[1pt/1pt] \addlinespace[2pt]
    $\text{BLOOM}_\text{7.1B}$     & 46.0   & 40.2   & 41.1   &  & 34.9 & 35.4 & 38.6 & 37.5 \\
     \ \ \ \ \ \ \ \ \ +MSFT             & 47.1   & 42.1   & 42.9   &  & 36.8 & 38.2 & 41.1 & 39.7 \\
     \ \ \ \ \ \ \ \ \ +AFP              & 47.9   & 43.4   & 43.6   &  & 37.9 & 39.3 & 41.8 & 40.6 \\
     \ \ \ \ \ \ \ \ \ +\textit{ShifCon} & 48.3   & 43.2   & 45.0   &  & 39.3 & 40.7 & 41.8 & 41.5 \\ \bottomrule[1.2pt]  
    \end{tabular}
    \caption{The detailed results of each language on the XNLI task in $\text{Llama-2}_\text{7B}$, $\text{XGLM}_\text{7.5B}$, and $\text{BLOOM}_\text{7.1B}$. 
    High- and low-resource languages are categorized based on their data ratios in the pre-training corpus.}
    \end{table*}

%% file: tabs/appendix_xstorycloze.tex
\begin{table*}[!th]
    \renewcommand\arraystretch{1.1}
    
    \centering
    \setlength\tabcolsep{4pt}
    \fontsize{10}{12}\selectfont 
    \begin{tabular}{lcccclcc}
    \toprule[1.2pt]
                     & \multicolumn{4}{c}{High}  &  & \multicolumn{2}{c}{Low} \\ \addlinespace[2pt]\cline{2-5} \cline{7-8} \addlinespace[2pt]
                     & EN   & ES   & ZH   & ID   &  & HI         & SW         \\ \hline \addlinespace[2pt]
    $\text{Llama-2}_\text{7B}$       & 84.4 & 75.5 & 69.4 & 69.4 &  & 57.9       & 55.3       \\
     \ \ \ \ \ \ \ \ \ +MSFT             & 85.5 & 76.9 & 70.5 & 68.3 &  & 59.6       & 57.8       \\
     \ \ \ \ \ \ \ \ \ +AFP              & 86.4 & 77.3 & 71.6 & 69.2 &  & 60.5       & 59.2       \\
     \ \ \ \ \ \ \ \ \ +\textit{ShifCon} & 86.2 & 77.5 & 72.8 & 70.1 &  & 60.2       & 61.5       \\ \addlinespace[2pt] \hdashline[1pt/1pt] \addlinespace[2pt]
    $\text{XGLM}_\text{7.5B}$      & 73.5 & 63.7 & 60.4 & 63.2 &  & 59.5       & 57.2       \\
     \ \ \ \ \ \ \ \ \ +MSFT             & 74.4 & 65.8 & 62.8 & 64.0 &  & 61.2       & 59.1       \\
     \ \ \ \ \ \ \ \ \ +AFP              & 75.5 & 66.7 & 63.0 & 65.1 &  & 61.9       & 60.0       \\
     \ \ \ \ \ \ \ \ \ +\textit{ShifCon} & 75.2 & 67.4 & 62.4 & 67.2 &  & 62.8       & 61.5       \\ \addlinespace[2pt] \hdashline[1pt/1pt] \addlinespace[2pt]
    $\text{BLOOM}_\text{7.1B}$     & 72.2 & 66.3 & 66.2 & 64.7 &  & 60.4       & 55.8       \\
     \ \ \ \ \ \ \ \ \ +MSFT             & 72.8 & 66.8 & 67.1 & 67.5 &  & 61.6       & 58.0       \\
     \ \ \ \ \ \ \ \ \ +AFP              & 72.2 & 67.5 & 67.9 & 67.2 &  & 61.9       & 58.5       \\
     \ \ \ \ \ \ \ \ \ +\textit{ShifCon} & 72.6 & 68.2 & 68.5 & 68.8 &  & 62.8       & 59.1       \\ \bottomrule[1.2pt]  
    \end{tabular}
    \caption{The detailed results of each language on the XStoryCloze task in $\text{Llama-2}_\text{7B}$, $\text{XGLM}_\text{7.5B}$, and $\text{BLOOM}_\text{7.1B}$. 
    High- and low-resource languages are categorized based on their data ratios in the pre-training corpus.}
    \end{table*}

%% file: appendix/appendix_lsd_all.tex
\input{tabs/appendix_lsd_all}

%% file: tabs/appendix_lsd_all.tex
\begin{table*}[!h]
\normalsize
\centering
\setlength \tabcolsep{4.5pt}
\begin{tabular}{lcc}
\toprule[1.2pt]
           & \textbf{Low Subspace Distance Area}  & \textbf{Layers} \\  \addlinespace[2pt]\cline{2-3}   \addlinespace[2pt]
$\text{Llama-2}_{\text{7B}}$ & [11, 20]              & 32              \\ 
$\text{Llama-3}_{\text{8B}}$ & [11, 20]              & 32              \\ \addlinespace[2pt] \hdashline[1pt/1pt] \addlinespace[2pt]
$\text{BLOOM}_{\text{7.1B}}$ & [14, 22]              & 30              \\
$\text{BLOOM}_{\text{1.7B}}$ & [10, 17]              & 24              \\
$\text{BLOOM}_{\text{560M}}$ & [10, 17]              & 24              \\ \addlinespace[2pt] \hdashline[1pt/1pt] \addlinespace[2pt]
$\text{XGLM}_{\text{7.5B}}$  & [13, 22]             & 32              \\
$\text{XGLM}_{\text{2.9B}}$  & [9, 23]              & 48              \\
$\text{XGLM}_{\text{564M}}$  & [9, 16]              & 24              \\ \bottomrule[1.2pt]
\end{tabular}
\caption{The low subspace distance areas of models in our experiments.}
\label{tab:all_lsd}
\end{table*}

%% file: appendix/appendix_language_code.tex
\begin{table*}[!h]
    \setlength{\tabcolsep}{2mm}
    \centering
    \renewcommand\arraystretch{1.25}
        \begin{tabular}{ccc}
            \toprule[1.2pt]  
            \multicolumn{1}{c}{\textbf{ISO 639-1}} & \multicolumn{1}{c}{\textbf{Language}}    &\multicolumn{1}{c}{\textbf{Family}} \\
                \midrule[0.8pt]
                BN      &Bengali      &Indo-European\\
                DE      &German      &Indo-European\\
                EN      &English      &Indo-European\\
                ES      &Spanish      &Indo-European\\
                FR      &French      &Indo-European\\
                HI      &Hindi      &Indo-European\\
                ID      &Indonesian      &Austronesian\\
                JA      &Japanese      &Japonic\\
                RU      &Russian      &Indo-European\\
                ZH      &Chinese      &Sino-Tibetan\\
                TH      &Thai      &Kra-Dai\\
                SW      &Swahili      &Niger-Congo\\
                TR      &Turkish      &Turkic\\
            \bottomrule[1.2pt]
        \end{tabular}
        \caption{\label{tab:lang_codes} Details of Language codes in this work.}
\end{table*}

%% file: acl_latex.bbl
\begin{thebibliography}{59}
\expandafter\ifx\csname natexlab\endcsname\relax\def\natexlab#1{#1}\fi

\bibitem[{Achiam et~al.(2023)Achiam, Adler, Agarwal, Ahmad, Akkaya, Aleman,
  Almeida, Altenschmidt, Altman, Anadkat et~al.}]{achiam2023gpt}
Josh Achiam, Steven Adler, Sandhini Agarwal, Lama Ahmad, Ilge Akkaya,
  Florencia~Leoni Aleman, Diogo Almeida, Janko Altenschmidt, Sam Altman,
  Shyamal Anadkat, et~al. 2023.
\newblock Gpt-4 technical report.
\newblock \emph{arXiv preprint arXiv:2303.08774}.

\bibitem[{Agrawal et~al.(2024)Agrawal, Fazili, and
  Jyothi}]{agrawal-etal-2024-translation}
Ashish Agrawal, Barah Fazili, and Preethi Jyothi. 2024.
\newblock \href {https://aclanthology.org/2024.eacl-short.28} {Translation
  errors significantly impact low-resource languages in cross-lingual
  learning}.
\newblock In \emph{Proceedings of the 18th Conference of the European Chapter
  of the Association for Computational Linguistics (Volume 2: Short Papers)},
  pages 319--329, St. Julian{'}s, Malta. Association for Computational
  Linguistics.

\bibitem[{Anil et~al.(2023)Anil, Dai, Firat, Johnson, Lepikhin, Passos,
  Shakeri, Taropa, Bailey, Chen et~al.}]{anil2023palm2}
Rohan Anil, Andrew~M Dai, Orhan Firat, Melvin Johnson, Dmitry Lepikhin,
  Alexandre Passos, Siamak Shakeri, Emanuel Taropa, Paige Bailey, Zhifeng Chen,
  et~al. 2023.
\newblock Palm 2 technical report.
\newblock \emph{arXiv preprint arXiv:2305.10403}.

\bibitem[{Bonnabel and Sepulchre(2009)}]{bonnabel-sepulchre-2009-riemannian}
Silv{\'e}re Bonnabel and Rodolphe Sepulchre. 2009.
\newblock \href {https://arxiv.org/abs/0807.4462} {Riemannian metric and
  geometric mean for positive semidefinite matrices of fixed rank}.
\newblock \emph{SIAM Journal on Matrix Analysis and Applications},
  31:1055--1070.

\bibitem[{Chang et~al.(2022)Chang, Tu, and Bergen}]{chang-etal-2022-geometry}
Tyler Chang, Zhuowen Tu, and Benjamin Bergen. 2022.
\newblock \href {https://doi.org/10.18653/v1/2022.emnlp-main.9} {The geometry
  of multilingual language model representations}.
\newblock In \emph{Proceedings of the 2022 Conference on Empirical Methods in
  Natural Language Processing}, pages 119--136, Abu Dhabi, United Arab
  Emirates. Association for Computational Linguistics.

\bibitem[{Chen et~al.(2023{\natexlab{a}})Chen, Zheng, Wu, Shou, Gong, Song,
  Zhang, and Li}]{chen2023breaking}
Nuo Chen, Zinan Zheng, Ning Wu, Linjun Shou, Ming Gong, Yangqiu Song, Dongmei
  Zhang, and Jia Li. 2023{\natexlab{a}}.
\newblock Breaking language barriers in multilingual mathematical reasoning:
  Insights and observations.
\newblock \emph{arXiv preprint arXiv:2310.20246}.

\bibitem[{Chen et~al.(2023{\natexlab{b}})Chen, Zheng, Wu, Shou, Gong, Song,
  Zhang, and Li}]{MathOctopus}
Nuo Chen, Zinan Zheng, Ning Wu, Linjun Shou, Ming Gong, Yangqiu Song, Dongmei
  Zhang, and Jia Li. 2023{\natexlab{b}}.
\newblock Breaking language barriers in multilingual mathematical reasoning:
  Insights and observations.
\newblock \emph{arXiv preprint arXiv:2310.20246}.

\bibitem[{Conneau et~al.(2018)Conneau, Rinott, Lample, Williams, Bowman,
  Schwenk, and Stoyanov}]{conneau-etal-2018-xnli}
Alexis Conneau, Ruty Rinott, Guillaume Lample, Adina Williams, Samuel Bowman,
  Holger Schwenk, and Veselin Stoyanov. 2018.
\newblock \href {https://doi.org/10.18653/v1/D18-1269} {{XNLI}: Evaluating
  cross-lingual sentence representations}.
\newblock In \emph{Proceedings of the 2018 Conference on Empirical Methods in
  Natural Language Processing}, pages 2475--2485, Brussels, Belgium.
  Association for Computational Linguistics.

\bibitem[{Conover et~al.(2023)Conover, Hayes, Mathur, Xie, Wan, Shah, Ghodsi,
  Wendell, Zaharia, and Xin}]{DatabricksBlog2023dolly}
Mike Conover, Matt Hayes, Ankit Mathur, Jianwei Xie, Jun Wan, Sam Shah, Ali
  Ghodsi, Patrick Wendell, Matei Zaharia, and Reynold Xin. 2023.
\newblock \href
  {https://www.databricks.com/blog/2023/04/12/dolly-first-open-commercially-viable-instruction-tuned-llm}
  {Free dolly: Introducing the world's first truly open instruction-tuned llm}.
\newblock \emph{databricks}.

\bibitem[{Grattafiori et~al.(2024)Grattafiori, Dubey, Jauhri, Pandey, Kadian,
  Al-Dahle, Letman, Mathur, Schelten, Vaughan
  et~al.}]{grattafiori2024llama3herdmodels}
Aaron Grattafiori, Abhimanyu Dubey, Abhinav Jauhri, Abhinav Pandey, Abhishek
  Kadian, Ahmad Al-Dahle, Aiesha Letman, Akhil Mathur, Alan Schelten, Alex
  Vaughan, et~al. 2024.
\newblock The llama 3 herd of models.
\newblock \emph{arXiv e-prints}, pages arXiv--2407.

\bibitem[{Gurgurov et~al.(2024)Gurgurov, B{\"a}umel, and
  Anikina}]{gurgurov2024multilingual}
Daniil Gurgurov, Tanja B{\"a}umel, and Tatiana Anikina. 2024.
\newblock Multilingual large language models and curse of multilinguality.
\newblock \emph{arXiv preprint arXiv:2406.10602}.

\bibitem[{Huang et~al.(2023)Huang, Tang, Zhang, Zhao, Song, Xia, and
  Wei}]{huang2023not}
Haoyang Huang, Tianyi Tang, Dongdong Zhang, Wayne~Xin Zhao, Ting Song, Yan Xia,
  and Furu Wei. 2023.
\newblock Not all languages are created equal in llms: Improving multilingual
  capability by cross-lingual-thought prompting.
\newblock In \emph{Findings of the Association for Computational Linguistics:
  EMNLP 2023}.

\bibitem[{Kassner et~al.(2021)Kassner, Dufter, and
  Sch{\"u}tze}]{kassner-etal-2021-multilingual}
Nora Kassner, Philipp Dufter, and Hinrich Sch{\"u}tze. 2021.
\newblock \href {https://doi.org/10.18653/v1/2021.eacl-main.284} {Multilingual
  {LAMA}: Investigating knowledge in multilingual pretrained language models}.
\newblock In \emph{Proceedings of the 16th Conference of the European Chapter
  of the Association for Computational Linguistics: Main Volume}, pages
  3250--3258, Online. Association for Computational Linguistics.

\bibitem[{Kholodna et~al.(2024)Kholodna, Julka, Khodadadi, Gumus, and
  Granitzer}]{kholodna2024llms}
Nataliia Kholodna, Sahib Julka, Mohammad Khodadadi, Muhammed~Nurullah Gumus,
  and Michael Granitzer. 2024.
\newblock Llms in the loop: Leveraging large language model annotations for
  active learning in low-resource languages.
\newblock In \emph{Joint European Conference on Machine Learning and Knowledge
  Discovery in Databases}, pages 397--412. Springer.

\bibitem[{Kojima et~al.(2024)Kojima, Okimura, Iwasawa, Yanaka, and
  Matsuo}]{kojima-etal-2024-multilingual}
Takeshi Kojima, Itsuki Okimura, Yusuke Iwasawa, Hitomi Yanaka, and Yutaka
  Matsuo. 2024.
\newblock \href {https://doi.org/10.18653/v1/2024.naacl-long.384} {On the
  multilingual ability of decoder-based pre-trained language models: Finding
  and controlling language-specific neurons}.
\newblock In \emph{Proceedings of the 2024 Conference of the North American
  Chapter of the Association for Computational Linguistics: Human Language
  Technologies (Volume 1: Long Papers)}, pages 6919--6971, Mexico City, Mexico.
  Association for Computational Linguistics.

\bibitem[{Kong et~al.(2022{\natexlab{a}})Kong, Chen, Zhang, Yang, and
  Yang}]{kong2022multitasking}
Cunliang Kong, Yun Chen, Hengyuan Zhang, Liner Yang, and Erhong Yang.
  2022{\natexlab{a}}.
\newblock Multitasking framework for unsupervised simple definition generation.
\newblock \emph{arXiv preprint arXiv:2203.12926}.

\bibitem[{Kong et~al.(2022{\natexlab{b}})Kong, Wang, Chong, Yang, Zhang, Yang,
  and Huang}]{kong2022blcu}
Cunliang Kong, Yujie Wang, Ruining Chong, Liner Yang, Hengyuan Zhang, Erhong
  Yang, and Yaping Huang. 2022{\natexlab{b}}.
\newblock Blcu-icall at semeval-2022 task 1: Cross-attention multitasking
  framework for definition modeling.
\newblock \emph{arXiv preprint arXiv:2204.07701}.

\bibitem[{Kozhevnikov and Titov(2014)}]{kozhevnikov2014cross}
Mikhail Kozhevnikov and Ivan Titov. 2014.
\newblock Cross-lingual model transfer using feature representation projection.
\newblock In \emph{Proceedings of the 52nd Annual Meeting of the Association
  for Computational Linguistics (Volume 2: Short Papers)}, pages 579--585.

\bibitem[{Li et~al.(2024{\natexlab{a}})Li, Wang, Zhang, and
  Zong}]{li-etal-2024-improving-context}
Chong Li, Shaonan Wang, Jiajun Zhang, and Chengqing Zong. 2024{\natexlab{a}}.
\newblock \href {https://doi.org/10.18653/v1/2024.naacl-long.445} {Improving
  in-context learning of multilingual generative language models with
  cross-lingual alignment}.
\newblock In \emph{Proceedings of the 2024 Conference of the North American
  Chapter of the Association for Computational Linguistics: Human Language
  Technologies (Volume 1: Long Papers)}, pages 8058--8076, Mexico City, Mexico.
  Association for Computational Linguistics.

\bibitem[{Li et~al.(2024{\natexlab{b}})Li, Tan, Chen, and
  Liu}]{li2024contextualization}
Dawei Li, Zhen Tan, Tianlong Chen, and Huan Liu. 2024{\natexlab{b}}.
\newblock Contextualization distillation from large language model for
  knowledge graph completion.
\newblock \emph{arXiv preprint arXiv:2402.01729}.

\bibitem[{Li et~al.(2024{\natexlab{c}})Li, Yang, Tan, Baik, Yun, Lee, Chacko,
  Hou, Duong-Tran, Ding et~al.}]{li2024dalk}
Dawei Li, Shu Yang, Zhen Tan, Jae~Young Baik, Sunkwon Yun, Joseph Lee, Aaron
  Chacko, Bojian Hou, Duy Duong-Tran, Ying Ding, et~al. 2024{\natexlab{c}}.
\newblock Dalk: Dynamic co-augmentation of llms and kg to answer alzheimer's
  disease questions with scientific literature.
\newblock \emph{arXiv preprint arXiv:2405.04819}.

\bibitem[{Li et~al.(2023{\natexlab{a}})Li, Zhang, Li, and Yang}]{li2023multi}
Dawei Li, Hengyuan Zhang, Yanran Li, and Shiping Yang. 2023{\natexlab{a}}.
\newblock Multi-level contrastive learning for script-based character
  understanding.
\newblock \emph{arXiv preprint arXiv:2310.13231}.

\bibitem[{Li et~al.(2023{\natexlab{b}})Li, Koto, Wu, Aji, and
  Baldwin}]{li2023bactrian}
Haonan Li, Fajri Koto, Minghao Wu, Alham~Fikri Aji, and Timothy Baldwin.
  2023{\natexlab{b}}.
\newblock Bactrian-x: Multilingual replicable instruction-following models with
  low-rank adaptation.
\newblock \emph{arXiv preprint arXiv:2305.15011}.

\bibitem[{Li and Murray(2023)}]{li-murray-2023-zero}
Tianjian Li and Kenton Murray. 2023.
\newblock \href {https://doi.org/10.18653/v1/2023.findings-acl.789} {Why does
  zero-shot cross-lingual generation fail? an explanation and a solution}.
\newblock In \emph{Findings of the Association for Computational Linguistics:
  ACL 2023}, pages 12461--12476, Toronto, Canada. Association for Computational
  Linguistics.

\bibitem[{Libovick{\'y} et~al.(2020)Libovick{\'y}, Rosa, and
  Fraser}]{libovicky-etal-2020-language}
Jind{\v{r}}ich Libovick{\'y}, Rudolf Rosa, and Alexander Fraser. 2020.
\newblock \href {https://doi.org/10.18653/v1/2020.findings-emnlp.150} {On the
  language neutrality of pre-trained multilingual representations}.
\newblock In \emph{Findings of the Association for Computational Linguistics:
  EMNLP 2020}, pages 1663--1674, Online. Association for Computational
  Linguistics.

\bibitem[{Lin et~al.(2022)Lin, Mihaylov, Artetxe, Wang, Chen, Simig, Ott,
  Goyal, Bhosale, Du, Pasunuru, Shleifer, Koura, Chaudhary, O{'}Horo, Wang,
  Zettlemoyer, Kozareva, Diab, Stoyanov, and Li}]{lin-etal-2022-shot}
Xi~Victoria Lin, Todor Mihaylov, Mikel Artetxe, Tianlu Wang, Shuohui Chen,
  Daniel Simig, Myle Ott, Naman Goyal, Shruti Bhosale, Jingfei Du, Ramakanth
  Pasunuru, Sam Shleifer, Punit~Singh Koura, Vishrav Chaudhary, Brian O{'}Horo,
  Jeff Wang, Luke Zettlemoyer, Zornitsa Kozareva, Mona Diab, Veselin Stoyanov,
  and Xian Li. 2022.
\newblock \href {https://doi.org/10.18653/v1/2022.emnlp-main.616} {Few-shot
  learning with multilingual generative language models}.
\newblock In \emph{Proceedings of the 2022 Conference on Empirical Methods in
  Natural Language Processing}, pages 9019--9052, Abu Dhabi, United Arab
  Emirates. Association for Computational Linguistics.

\bibitem[{Liu et~al.(2024)Liu, Ma, Ye, and Schuetze}]{liu-etal-2024-translico}
Yihong Liu, Chunlan Ma, Haotian Ye, and Hinrich Schuetze. 2024.
\newblock \href {https://aclanthology.org/2024.acl-long.136} {{T}ransli{C}o: A
  contrastive learning framework to address the script barrier in multilingual
  pretrained language models}.
\newblock In \emph{Proceedings of the 62nd Annual Meeting of the Association
  for Computational Linguistics (Volume 1: Long Papers)}, pages 2476--2499,
  Bangkok, Thailand. Association for Computational Linguistics.

\bibitem[{Loshchilov and Hutter(2019)}]{loshchilov2018decoupled}
Ilya Loshchilov and Frank Hutter. 2019.
\newblock \href {https://openreview.net/forum?id=Bkg6RiCqY7} {Decoupled weight
  decay regularization}.
\newblock In \emph{International Conference on Learning Representations}.

\bibitem[{Muennighoff et~al.(2023)Muennighoff, Wang, Sutawika, Roberts,
  Biderman, Le~Scao, Bari, Shen, Yong, Schoelkopf, Tang, Radev, Aji, Almubarak,
  Albanie, Alyafeai, Webson, Raff, and
  Raffel}]{muennighoff-etal-2023-crosslingual}
Niklas Muennighoff, Thomas Wang, Lintang Sutawika, Adam Roberts, Stella
  Biderman, Teven Le~Scao, M~Saiful Bari, Sheng Shen, Zheng~Xin Yong, Hailey
  Schoelkopf, Xiangru Tang, Dragomir Radev, Alham~Fikri Aji, Khalid Almubarak,
  Samuel Albanie, Zaid Alyafeai, Albert Webson, Edward Raff, and Colin Raffel.
  2023.
\newblock \href {https://doi.org/10.18653/v1/2023.acl-long.891} {Crosslingual
  generalization through multitask finetuning}.
\newblock In \emph{Proceedings of the 61st Annual Meeting of the Association
  for Computational Linguistics (Volume 1: Long Papers)}, pages 15991--16111,
  Toronto, Canada. Association for Computational Linguistics.

\bibitem[{Ponti et~al.(2020)Ponti, Glava{\v{s}}, Majewska, Liu, Vuli{\'c}, and
  Korhonen}]{ponti-etal-2020-xcopa}
Edoardo~Maria Ponti, Goran Glava{\v{s}}, Olga Majewska, Qianchu Liu, Ivan
  Vuli{\'c}, and Anna Korhonen. 2020.
\newblock \href {https://doi.org/10.18653/v1/2020.emnlp-main.185} {{XCOPA}: A
  multilingual dataset for causal commonsense reasoning}.
\newblock In \emph{Proceedings of the 2020 Conference on Empirical Methods in
  Natural Language Processing (EMNLP)}, pages 2362--2376, Online. Association
  for Computational Linguistics.

\bibitem[{Popovi{\'c}(2017)}]{popovic-2017-chrf}
Maja Popovi{\'c}. 2017.
\newblock \href {https://doi.org/10.18653/v1/W17-4770} {chr{F}++: words helping
  character n-grams}.
\newblock In \emph{Proceedings of the Second Conference on Machine
  Translation}, pages 612--618, Copenhagen, Denmark. Association for
  Computational Linguistics.

\bibitem[{Radford et~al.(2021)Radford, Kim, Hallacy, Ramesh, Goh, Agarwal,
  Sastry, Askell, Mishkin, Clark et~al.}]{radford2021learning}
Alec Radford, Jong~Wook Kim, Chris Hallacy, Aditya Ramesh, Gabriel Goh,
  Sandhini Agarwal, Girish Sastry, Amanda Askell, Pamela Mishkin, Jack Clark,
  et~al. 2021.
\newblock Learning transferable visual models from natural language
  supervision.
\newblock In \emph{International Conference on Machine Learning}, pages
  8748--8763. PMLR.

\bibitem[{Scao et~al.(2022)Scao, Fan, Akiki, Pavlick, Ili{\'c}, Hesslow,
  Castagn{\'e}, Luccioni, Yvon, Gall{\'e} et~al.}]{scao2022bloom}
Teven~Le Scao, Angela Fan, Christopher Akiki, Ellie Pavlick, Suzana Ili{\'c},
  Daniel Hesslow, Roman Castagn{\'e}, Alexandra~Sasha Luccioni, Fran{\c{c}}ois
  Yvon, Matthias Gall{\'e}, et~al. 2022.
\newblock \href {https://arxiv.org/abs/2211.05100} {Bloom: A 176b-parameter
  open-access multilingual language model}.
\newblock \emph{arXiv preprint arXiv:2211.05100}.

\bibitem[{Shi et~al.()Shi, Suzgun, Freitag, Wang, Srivats, Vosoughi, Chung,
  Tay, Ruder, Zhou et~al.}]{shilanguage}
Freda Shi, Mirac Suzgun, Markus Freitag, Xuezhi Wang, Suraj Srivats, Soroush
  Vosoughi, Hyung~Won Chung, Yi~Tay, Sebastian Ruder, Denny Zhou, et~al.
\newblock Language models are multilingual chain-of-thought reasoners.
\newblock In \emph{The Eleventh International Conference on Learning
  Representations}.

\bibitem[{Shi et~al.(2022)Shi, Suzgun, Freitag, Wang, Srivats, Vosoughi, Chung,
  Tay, Ruder, Zhou et~al.}]{shi2022language}
Freda Shi, Mirac Suzgun, Markus Freitag, Xuezhi Wang, Suraj Srivats, Soroush
  Vosoughi, Hyung~Won Chung, Yi~Tay, Sebastian Ruder, Denny Zhou, et~al. 2022.
\newblock Language models are multilingual chain-of-thought reasoners.
\newblock In \emph{International Conference on Learning Representations
  (ICLR)}.

\bibitem[{Tan et~al.(2024)Tan, Beigi, Wang, Guo, Bhattacharjee, Jiang, Karami,
  Li, Cheng, and Liu}]{tan2024large}
Zhen Tan, Alimohammad Beigi, Song Wang, Ruocheng Guo, Amrita Bhattacharjee,
  Bohan Jiang, Mansooreh Karami, Jundong Li, Lu~Cheng, and Huan Liu. 2024.
\newblock Large language models for data annotation: A survey.
\newblock \emph{arXiv preprint arXiv:2402.13446}.

\bibitem[{Tang et~al.(2024)Tang, Luo, Huang, Zhang, Wang, Zhao, Wei, and
  Wen}]{tang-etal-2024-language}
Tianyi Tang, Wenyang Luo, Haoyang Huang, Dongdong Zhang, Xiaolei Wang, Xin
  Zhao, Furu Wei, and Ji-Rong Wen. 2024.
\newblock \href {https://aclanthology.org/2024.acl-long.309} {Language-specific
  neurons: The key to multilingual capabilities in large language models}.
\newblock In \emph{Proceedings of the 62nd Annual Meeting of the Association
  for Computational Linguistics (Volume 1: Long Papers)}, pages 5701--5715,
  Bangkok, Thailand. Association for Computational Linguistics.

\bibitem[{Taori et~al.(2023)Taori, Gulrajani, Zhang, Dubois, Li, Guestrin,
  Liang, and Hashimoto}]{alpaca}
Rohan Taori, Ishaan Gulrajani, Tianyi Zhang, Yann Dubois, Xuechen Li, Carlos
  Guestrin, Percy Liang, and Tatsunori~B. Hashimoto. 2023.
\newblock Stanford alpaca: An instruction-following llama model.
\newblock \url{https://github.com/tatsu-lab/stanford_alpaca}.

\bibitem[{Team"(2022)}]{nllb2022}
"NLLB Team". 2022.
\newblock No language left behind: Scaling human-centered machine translation.

\bibitem[{Tong et~al.(2024)Tong, Li, Wang, Wang, Teng, and Shang}]{tong2024can}
Yongqi Tong, Dawei Li, Sizhe Wang, Yujia Wang, Fei Teng, and Jingbo Shang.
  2024.
\newblock Can llms learn from previous mistakes? investigating llms' errors to
  boost for reasoning.
\newblock \emph{arXiv preprint arXiv:2403.20046}.

\bibitem[{Touvron et~al.(2023)Touvron, Martin, Stone, Albert, Almahairi,
  Babaei, Bashlykov, Batra, Bhargava, Bhosale, Bikel, Blecher, Ferrer, Chen,
  Cucurull, Esiobu, Fernandes, Fu, Fu, Fuller, Gao, Goswami, Goyal, Hartshorn,
  Hosseini, Hou, Inan, Kardas, Kerkez, Khabsa, Kloumann, Korenev, Koura,
  Lachaux, Lavril, Lee, Liskovich, Lu, Mao, Martinet, Mihaylov, Mishra,
  Molybog, Nie, Poulton, Reizenstein, Rungta, Saladi, Schelten, Silva, Smith,
  Subramanian, Tan, Tang, Taylor, Williams, Kuan, Xu, Yan, Zarov, Zhang, Fan,
  Kambadur, Narang, Rodriguez, Stojnic, Edunov, and
  Scialom}]{touvron2023llama2openfoundation}
Hugo Touvron, Louis Martin, Kevin Stone, Peter Albert, Amjad Almahairi, Yasmine
  Babaei, Nikolay Bashlykov, Soumya Batra, Prajjwal Bhargava, Shruti Bhosale,
  Dan Bikel, Lukas Blecher, Cristian~Canton Ferrer, Moya Chen, Guillem
  Cucurull, David Esiobu, Jude Fernandes, Jeremy Fu, Wenyin Fu, Brian Fuller,
  Cynthia Gao, Vedanuj Goswami, Naman Goyal, Anthony Hartshorn, Saghar
  Hosseini, Rui Hou, Hakan Inan, Marcin Kardas, Viktor Kerkez, Madian Khabsa,
  Isabel Kloumann, Artem Korenev, Punit~Singh Koura, Marie-Anne Lachaux,
  Thibaut Lavril, Jenya Lee, Diana Liskovich, Yinghai Lu, Yuning Mao, Xavier
  Martinet, Todor Mihaylov, Pushkar Mishra, Igor Molybog, Yixin Nie, Andrew
  Poulton, Jeremy Reizenstein, Rashi Rungta, Kalyan Saladi, Alan Schelten, Ruan
  Silva, Eric~Michael Smith, Ranjan Subramanian, Xiaoqing~Ellen Tan, Binh Tang,
  Ross Taylor, Adina Williams, Jian~Xiang Kuan, Puxin Xu, Zheng Yan, Iliyan
  Zarov, Yuchen Zhang, Angela Fan, Melanie Kambadur, Sharan Narang, Aurelien
  Rodriguez, Robert Stojnic, Sergey Edunov, and Thomas Scialom. 2023.
\newblock \href {http://arxiv.org/abs/2307.09288} {Llama 2: Open foundation and
  fine-tuned chat models}.

\bibitem[{Wang et~al.(2024)Wang, Tong, Zhang, Li, Zhang, and
  Chen}]{wang2024bpo}
Sizhe Wang, Yongqi Tong, Hengyuan Zhang, Dawei Li, Xin Zhang, and Tianlong
  Chen. 2024.
\newblock Bpo: Towards balanced preference optimization between knowledge
  breadth and depth in alignment.
\newblock \emph{arXiv preprint arXiv:2411.10914}.

\bibitem[{Wen-Yi and Mimno(2023)}]{wen-yi-mimno-2023-hyperpolyglot}
Andrea~W Wen-Yi and David Mimno. 2023.
\newblock \href {https://doi.org/10.18653/v1/2023.emnlp-main.71} {Hyperpolyglot
  {LLM}s: Cross-lingual interpretability in token embeddings}.
\newblock In \emph{Proceedings of the 2023 Conference on Empirical Methods in
  Natural Language Processing}, pages 1124--1131, Singapore. Association for
  Computational Linguistics.

\bibitem[{Xu et~al.(2023)Xu, Li, and
  Xiong}]{xu-etal-2023-language-representation}
Shaoyang Xu, Junzhuo Li, and Deyi Xiong. 2023.
\newblock \href {https://doi.org/10.18653/v1/2023.emnlp-main.226} {Language
  representation projection: Can we transfer factual knowledge across languages
  in multilingual language models?}
\newblock In \emph{Proceedings of the 2023 Conference on Empirical Methods in
  Natural Language Processing}, pages 3692--3702, Singapore. Association for
  Computational Linguistics.

\bibitem[{Xue et~al.(2021)Xue, Constant, Roberts, Kale, Al-Rfou, Siddhant,
  Barua, and Raffel}]{xue-etal-2021-mt5}
Linting Xue, Noah Constant, Adam Roberts, Mihir Kale, Rami Al-Rfou, Aditya
  Siddhant, Aditya Barua, and Colin Raffel. 2021.
\newblock \href {https://doi.org/10.18653/v1/2021.naacl-main.41} {m{T}5: A
  massively multilingual pre-trained text-to-text transformer}.
\newblock In \emph{Proceedings of the 2021 Conference of the North American
  Chapter of the Association for Computational Linguistics: Human Language
  Technologies}, pages 483--498, Online. Association for Computational
  Linguistics.

\bibitem[{Yao et~al.(2024)Yao, Zhuang, Sun, Xu, Kumar, and Shang}]{yao2024data}
Feng Yao, Yufan Zhuang, Zihao Sun, Sunan Xu, Animesh Kumar, and Jingbo Shang.
  2024.
\newblock Data contamination can cross language barriers.
\newblock \emph{arXiv preprint arXiv:2406.13236}.

\bibitem[{Yin et~al.(2022)Yin, Bansal, Monajatipoor, Li, and
  Chang}]{yin-etal-2022-geomlama}
Da~Yin, Hritik Bansal, Masoud Monajatipoor, Liunian~Harold Li, and Kai-Wei
  Chang. 2022.
\newblock \href {https://doi.org/10.18653/v1/2022.emnlp-main.132}
  {{G}eo{MLAMA}: Geo-diverse commonsense probing on multilingual pre-trained
  language models}.
\newblock In \emph{Proceedings of the 2022 Conference on Empirical Methods in
  Natural Language Processing}, pages 2039--2055, Abu Dhabi, United Arab
  Emirates. Association for Computational Linguistics.

\bibitem[{Yoon et~al.(2024)Yoon, Jang, Kim, Kim, Shafayat, and
  Seo}]{yoon-etal-2024-langbridge}
Dongkeun Yoon, Joel Jang, Sungdong Kim, Seungone Kim, Sheikh Shafayat, and
  Minjoon Seo. 2024.
\newblock \href {https://doi.org/10.18653/v1/2024.acl-long.405}
  {{L}ang{B}ridge: Multilingual reasoning without multilingual supervision}.
\newblock In \emph{Proceedings of the 62nd Annual Meeting of the Association
  for Computational Linguistics (Volume 1: Long Papers)}, pages 7502--7522,
  Bangkok, Thailand. Association for Computational Linguistics.

\bibitem[{Zhang et~al.(2025{\natexlab{a}})Zhang, Chen, Qiu, Liang, Li, Wang,
  Li, Mo, Li, So et~al.}]{zhang2025guilomo}
Hengyuan Zhang, Xinrong Chen, Yingmin Qiu, Xiao Liang, Ziyue Li, Guanyu Wang,
  Weiping Li, Tong Mo, Wenyue Li, Hayden Kwok-Hay So, et~al.
  2025{\natexlab{a}}.
\newblock Guilomo: Allocating expert number and rank for lora-moe via bilevel
  optimization with guidedselection vectors.
\newblock \emph{arXiv preprint arXiv:2506.14646}.

\bibitem[{Zhang et~al.(2023{\natexlab{a}})Zhang, Li, Li, Shang, Shi, and
  Jiang}]{zhang2023assisting}
Hengyuan Zhang, Dawei Li, Yanran Li, Chenming Shang, Chufan Shi, and Yong
  Jiang. 2023{\natexlab{a}}.
\newblock Assisting language learners: Automated trans-lingual definition
  generation via contrastive prompt learning.
\newblock \emph{arXiv preprint arXiv:2306.06058}.

\bibitem[{Zhang et~al.(2022)Zhang, Li, Yang, and Li}]{zhang2022fine}
Hengyuan Zhang, Dawei Li, Shiping Yang, and Yanran Li. 2022.
\newblock Fine-grained contrastive learning for definition generation.

\bibitem[{Zhang et~al.(2025{\natexlab{b}})Zhang, Liu, Shang, Li, and
  Jiang}]{zhang2024question}
Hengyuan Zhang, Zitao Liu, Chenming Shang, Dawei Li, and Yong Jiang.
  2025{\natexlab{b}}.
\newblock A question-centric multi-experts contrastive learning framework for
  improving the accuracy and interpretability of deep sequential knowledge
  tracing models.
\newblock \emph{ACM Transactions on Knowledge Discovery from Data},
  19(2):1--25.

\bibitem[{Zhang et~al.(2024{\natexlab{a}})Zhang, Wu, Li, Yang, Zhao, Jiang, and
  Tan}]{zhang-etal-2024-balancing}
Hengyuan Zhang, Yanru Wu, Dawei Li, Sak Yang, Rui Zhao, Yong Jiang, and Fei
  Tan. 2024{\natexlab{a}}.
\newblock Balancing speciality and versatility: a coarse to fine framework for
  supervised fine-tuning large language model.
\newblock In \emph{Findings of the Association for Computational Linguistics
  ACL 2024}, pages 7467--7509.

\bibitem[{Zhang et~al.(2024{\natexlab{b}})Zhang, Jin, Huang, Zhang, and
  Wei}]{zhang-etal-2024-respond}
Liang Zhang, Qin Jin, Haoyang Huang, Dongdong Zhang, and Furu Wei.
  2024{\natexlab{b}}.
\newblock \href {https://doi.org/10.18653/v1/2024.acl-long.229} {Respond in my
  language: Mitigating language inconsistency in response generation based on
  large language models}.
\newblock In \emph{Proceedings of the 62nd Annual Meeting of the Association
  for Computational Linguistics (Volume 1: Long Papers)}, pages 4177--4192,
  Bangkok, Thailand. Association for Computational Linguistics.

\bibitem[{Zhang et~al.(2024{\natexlab{c}})Zhang, Gautam, Wang, Alabi, Shen,
  Klakow, and Mosbach}]{zhang-etal-2024-impact}
Miaoran Zhang, Vagrant Gautam, Mingyang Wang, Jesujoba Alabi, Xiaoyu Shen,
  Dietrich Klakow, and Marius Mosbach. 2024{\natexlab{c}}.
\newblock \href {https://doi.org/10.18653/v1/2024.findings-acl.438} {The impact
  of demonstrations on multilingual in-context learning: A multidimensional
  analysis}.
\newblock In \emph{Findings of the Association for Computational Linguistics
  ACL 2024}, pages 7342--7371, Bangkok, Thailand and virtual meeting.
  Association for Computational Linguistics.

\bibitem[{Zhang et~al.(2023{\natexlab{b}})Zhang, Fang, Zhang, Ma, Zhou, Huang,
  Bu, Gui, Chen, Chen et~al.}]{zhang2023bayling}
Shaolei Zhang, Qingkai Fang, Zhuocheng Zhang, Zhengrui Ma, Yan Zhou, Langlin
  Huang, Mengyu Bu, Shangtong Gui, Yunji Chen, Xilin Chen, et~al.
  2023{\natexlab{b}}.
\newblock Bayling: Bridging cross-lingual alignment and instruction following
  through interactive translation for large language models.
\newblock \emph{arXiv preprint arXiv:2306.10968}.

\bibitem[{Zhao et~al.(2024)Zhao, Yoshinaga, and Oba}]{zhao-etal-2024-tracing}
Xin Zhao, Naoki Yoshinaga, and Daisuke Oba. 2024.
\newblock \href {https://aclanthology.org/2024.eacl-long.127} {Tracing the
  roots of facts in multilingual language models: Independent, shared, and
  transferred knowledge}.
\newblock In \emph{Proceedings of the 18th Conference of the European Chapter
  of the Association for Computational Linguistics (Volume 1: Long Papers)},
  pages 2088--2102, St. Julian{'}s, Malta. Association for Computational
  Linguistics.

\bibitem[{Zhu et~al.(2024{\natexlab{a}})Zhu, Wu, Bai, Song, and
  Gao}]{zhu2024eeg}
Mu~Zhu, Qingzhou Wu, Zhongli Bai, Yu~Song, and Qiang Gao. 2024{\natexlab{a}}.
\newblock Eeg-eye movement based subject dependence, cross-subject, and
  cross-session emotion recognition with multidimensional homogeneous encoding
  space alignment.
\newblock \emph{Expert Systems with Applications}, 251:124001.

\bibitem[{Zhu et~al.(2024{\natexlab{b}})Zhu, Liu, Dong, Xu, Huang, Kong, Chen,
  and Li}]{zhu2023multilingual}
Wenhao Zhu, Hongyi Liu, Qingxiu Dong, Jingjing Xu, Shujian Huang, Lingpeng
  Kong, Jiajun Chen, and Lei Li. 2024{\natexlab{b}}.
\newblock Multilingual machine translation with large language models:
  Empirical results and analysis.
\newblock In \emph{Findings of the Association for Computational Linguistics:
  NAACL 2024}.

\end{thebibliography}
